\documentclass[10pt,twocolumn,letterpaper]{article}

\usepackage{iccv}              

\usepackage[accsupp]{axessibility}  

\newcommand {\fredo}[1]{}
\newcommand {\caroline}[1]{}
\newcommand {\caro}[1]{}
\newcommand {\hyojin}[1]{}
\newcommand {\phillip}[1]{}
\newcommand {\todo}[1]{}

\usepackage{xcolor,colortbl,afterpage,capt-of,multicol}

\usepackage{amssymb}
\usepackage{pifont}

\usepackage{mathtools}
\newcommand{\defeq}{\vcentcolon=}

\definecolor{wino}{HTML}{cddefa}
\definecolor{dc4870}{HTML}{e0fccf}
\definecolor{genai}{HTML}{fae6ec}

\usepackage{siunitx}
\usepackage{numprint}

\definecolor{iccvblue}{rgb}{0.21,0.49,0.74}
\definecolor{myapricot}{RGB}{249,224,199}
\definecolor{mygray}{RGB}{238,238,238}
\usepackage[dvipsnames]{xcolor}
\usepackage[pagebackref,breaklinks,colorlinks,allcolors=iccvblue]
{hyperref}
\def\confYear{2025}

\title{Cycle Consistency as Reward: Learning Image-Text Alignment\\without Human Preferences}

\author{Hyojin Bahng\thanks{Equal contribution.}
\and
Caroline Chan\footnotemark[1]
\and
Fr\'edo Durand
\and
Phillip Isola
\and
MIT CSAIL\\
{\tt\small \{bahng, cmchan, fredo, phillipi\}@mit.edu}
}

\begin{document}
\maketitle

\begin{abstract}
Measuring alignment between language and vision is a fundamental challenge, especially as multimodal data becomes increasingly detailed and complex. Existing methods often rely on collecting human or AI preferences, which can be costly and time-intensive.
We propose an alternative approach that leverages cycle consistency as a supervisory signal. Given an image and generated text, we map the text back to image space using a text-to-image model and compute the similarity between the original image and its reconstruction. Analogously, for text-to-image generation, we measure the textual similarity between an input caption and its reconstruction through the cycle. We use the cycle consistency score to rank candidates and construct a preference dataset of 866K comparison pairs. The reward model trained on our dataset, CycleReward, outperforms state-of-the-art alignment metrics on detailed captioning, with superior inference-time scalability when used as a verifier for Best-of-$N$ sampling, while maintaining speed and differentiability. 
Furthermore, performing DPO and Diffusion DPO using our dataset enhances performance across a wide range of vision-language tasks and text-to-image generation. Our dataset, model, and code are publicly released at \href{https://cyclereward.github.io/}{https://cyclereward.github.io/}.
\end{abstract}

\begin{figure*}[!h]
  \centering
   \includegraphics[width=0.96\linewidth]{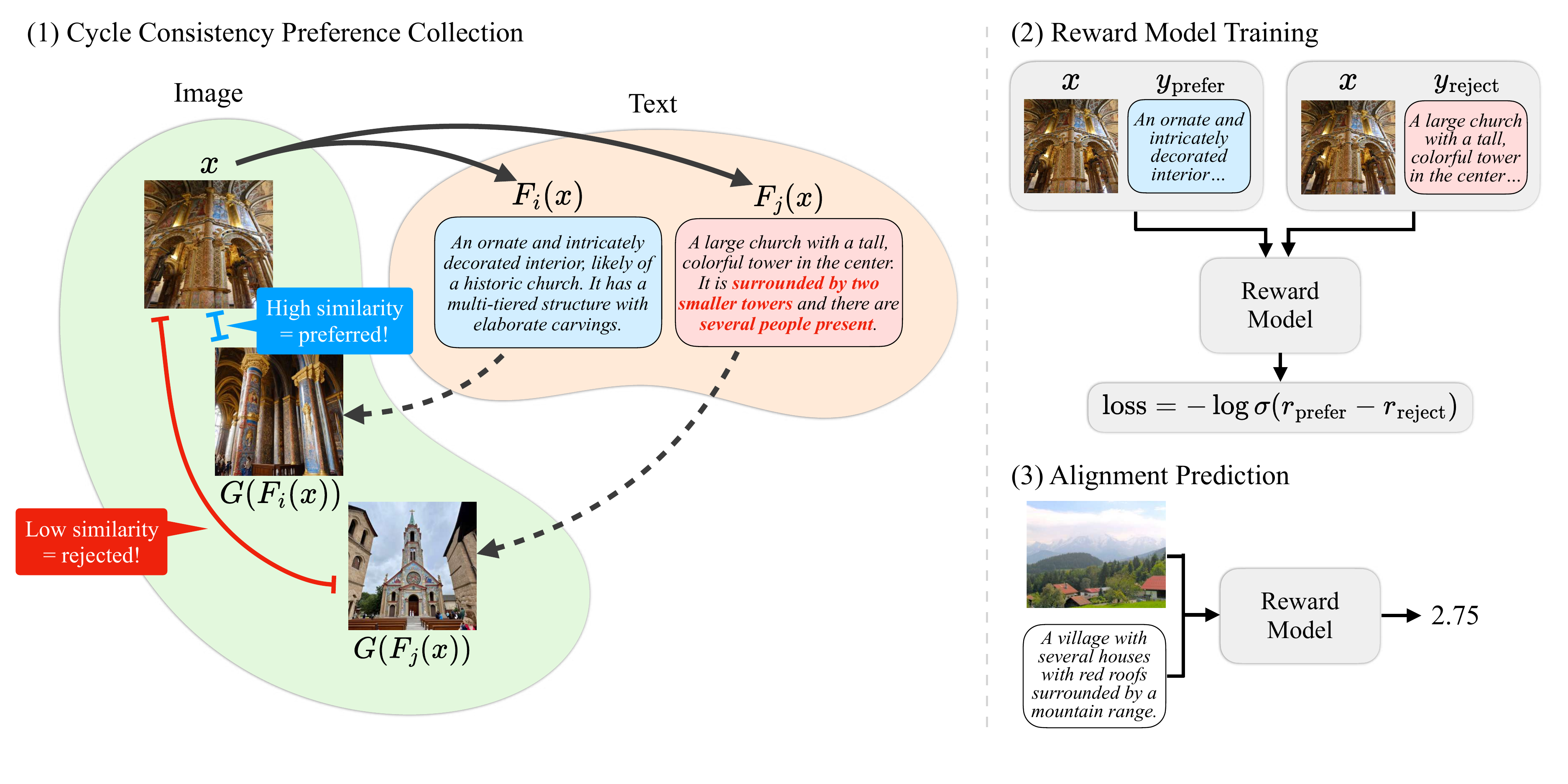}
   \vspace{-0.6cm}
   \caption{\textbf{Method overview.} 
   (1) Given an input image $x$, we generate multiple candidate captions $F_i(x)$, $F_j(x)$ using different captioning models. Each caption is mapped back to the image domain via a text-to-image model $G$, and compared against the original image. Captions whose reconstructions $G(F(x))$ are more similar to the original image are preferred; those with low similarity are rejected. (2) These comparison pairs are used to train a reward model, which learns to assign higher scores to preferred captions. We apply the same process for text-to-image generation. (3) At test time, the trained reward model outputs alignment scores for arbitrary image-text pairs.
   }
   \label{fig:overview}
   \vspace{-0.3cm}
\end{figure*}

\section{Introduction}

Measuring image–text alignment is a central problem in multimodal learning, where the goal is to learn a metric $d(x,y)$ that quantifies the correspondence between an image $x$ and text $y$. Such metrics are essential for evaluating vision–language and text-to-image models~\cite{hessel2021clipscore, lin2024evaluating, xu2024imagereward, kirstain2023pick, wu2023human} and improving model alignment through test-time optimization~\cite{cobbe2021training,stiennon2020learning,ma2025inference} or reinforcement learning from human feedback (RLHF)~\cite{ouyang2022training}. However, existing metrics typically rely on high-quality human preference data~\cite{xu2024imagereward, kirstain2023pick, wu2023human, wu2023humanv2}, which are expensive to collect and difficult to scale. Moreover, most of these datasets focus on short text~\cite{xu2024imagereward, kirstain2023pick, wu2023human}, limiting their ability to assess alignment for longer and more complex text. Another method uses AI feedback~\cite{silkie} from proprietary models (e.g., GPT-4V~\cite{gpt-4v}), which are costly, closed-source, and rate-limited via APIs, limiting long-term accessibility and scalability.

Comparing images and text is inherently challenging, especially with longer, detailed text. However, the comparison becomes much easier when we map text back into image space.
As shown in Figure~\ref{fig:overview}, more descriptive and accurate texts lead to reconstructed images that better resemble the original images.
This idea of cycle consistency~\cite{kalal2010forward,sundaram2010dense,zhu2017unpaired} has been used as a metric to evaluate image-to-text generation~\cite{huang2025image2text2image, garg2024imageinwords} and optimize diffusion models~\cite{blacktraining}. However, these approaches compute cycle consistency on-the-fly using large pre-trained models, which is prohibitively slow and often not differentiable.

We introduce CycleReward, a reward model trained on preferences derived from cycle consistency. Given an image-to-text mapping $F:X \rightarrow Y$ and a backward text-to-image mapping $G: Y \rightarrow X$, we define cycle consistency score as the similarity between the original input $x$ and its reconstruction $G(F(x))$. In the opposite direction, we can compare reconstructed text $F(G(y))$ with input text $y$. We use the cycle consistency score as a proxy for preferences, where a higher score indicates a preferred output. This provides a more scalable and cheaper signal for learning alignment compared to human supervision. We create a large-scale preference dataset, CyclePrefDB, comprising 866K comparison pairs from 11 image-to-text models and 4 text-to-image models. It contains significantly denser text than typical text-to-image datasets (Table~\ref{tab:data_comparison}), while fitting within the 77-token limit of text-to-image models. Trained on this dataset, CycleReward is a fast, differentiable metric for image-text alignment, particularly for longer text.

We evaluate CycleReward's ability to evaluate and enhance image-text alignment across two tasks: detailed captioning and text-to-image generation. We find that it is effective both metric for evaluation and Best-of-$N$ optimization. It achieves state-of-the-art performance for detailed captioning and performs competitively text-to-image synthesis. Finally, applying direct preference optimization (DPO)~\cite{rafailov2023direct,wallace2024diffusion} using CyclePrefDB enhances a wide range of vision-language and text-to-image generation tasks without requiring any human supervision.

\noindent
In summary, we make the following contributions:
\begin{itemize}

\item \textbf{CyclePrefDB}, a cycle consistency based preference dataset of 866K comparisons for image-to-text and text-to-image generation, specifically for \textit{longer} texts. 
\item \textbf{CycleReward}, a reward model trained on our dataset, which is effective as a fast, differentiable alignment metric and a verifier for Best-of-$N$ sampling for longer text. 
\item Our ablation study finds that image-to-text decoders with stronger language models lead to better alignment. Additionally, using similarity metrics that model human perception improves alignment.
\item Demonstration of DPO using our CyclePrefDB dataset, leading to improvements on a wide range of vision-language and text-to-image generation tasks. 

\end{itemize}

\begin{figure*}[t] 
  \centering
   \includegraphics[width=0.99\linewidth]{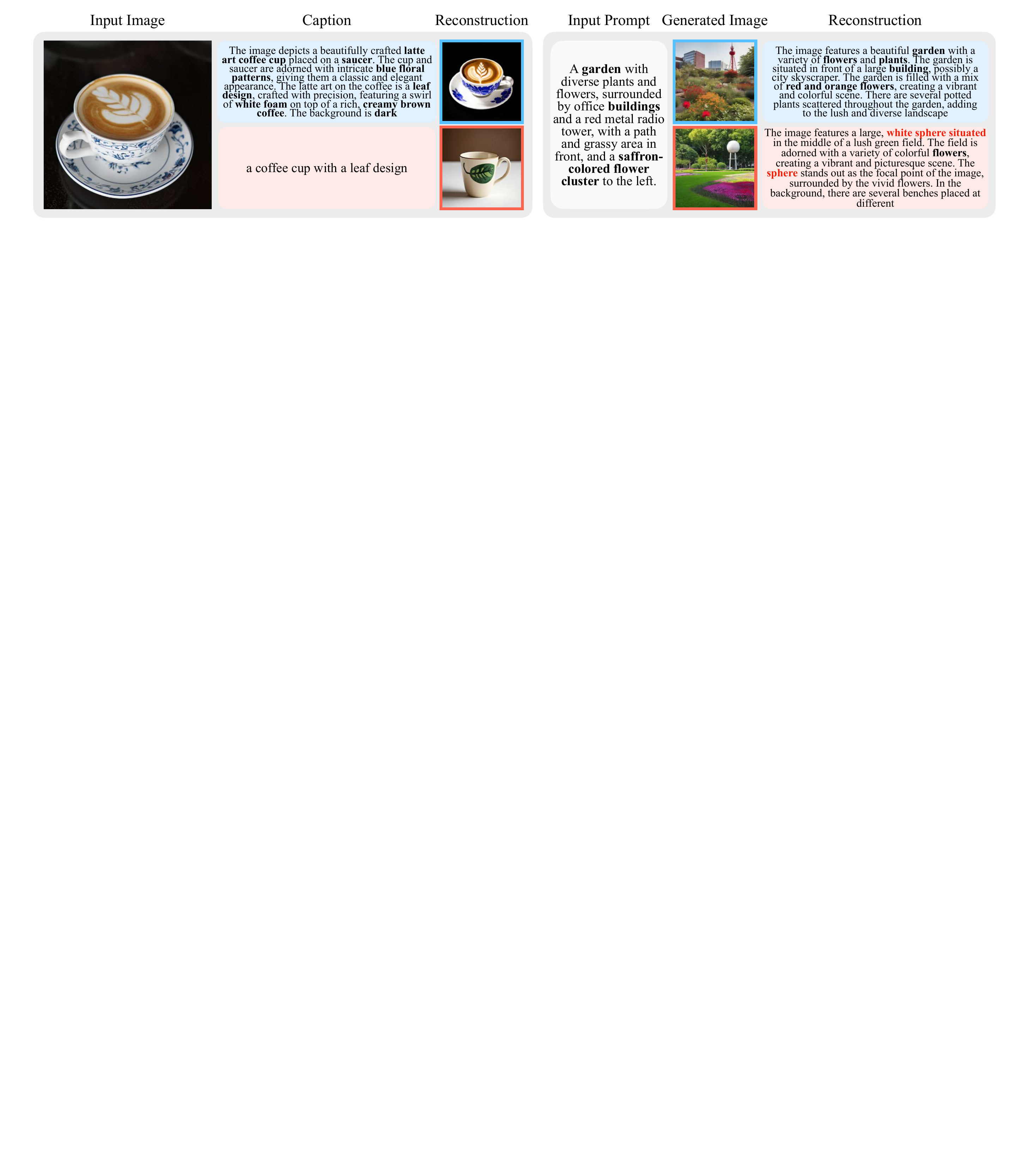}
   \vspace{-0.1cm}
   \caption{\textbf{What do cycle consistency preferences look like?} We visualize comparison pairs from our dataset, where cycle consistency determines preferences. \textcolor{cyan}{Preferred} samples are in blue and \textcolor{red}{rejected} samples are in red. \textbf{Image-to-text generation} (left): The preferred caption provides a fine-grained description resulting in a faithful reconstruction of the original image, whereas the rejected caption is short and vague, producing a reconstruction far from the original image. \textbf{Text-to-image generation} (right): Images that capture fine-grained details of the input prompt produce better text reconstructions, resulting in higher cycle consistency. See Appendix~\ref{appendix:dataset_recons} for more examples.}
   \label{fig:main_figure}
   \vspace{-0.3cm}
\end{figure*}

\section{Related Work}

\noindent
\textbf{Image-text alignment.}
Image-text alignment metrics can be classified either as reference-based, which require comparison with ground truth text, or as reference-free, which compute alignment based solely on the provided image and text. Reference-based metrics include BLEU~\cite{papineni2002bleu}, CIDEr~\cite{vedantam2015cider}, and METEOR~\cite{lavie2007meteor} which measure linguistic similarity between candidate and reference captions, but often do not generalize well to texts which vary in style and syntax from the reference caption. Recent approaches such as SPICE~\cite{anderson2016spice}, CAPTURE~\cite{dong2024benchmarking}, and DCScore~\cite{yepainting} decompose the candidate text into scene graphs or basic information units which are then compared to ground truth labels. Although these recent metrics are more flexible and thorough, they are limited by lack of differentiability, slow runtime, and, most importantly, they require a reference, which means they are not suitable as an objective function.

Reference-free metrics come in a variety of forms. Many approaches adapt pre-trained CLIP~\cite{radford2021learning} image and text encodings~\cite{hessel2021clipscore,doveh2023dense,jin2023refclip,yuksekgonuland} while others collect human preferences to train reward models~\cite{kirstain2023pick,xu2024imagereward,wu2023human,wu2023humanv2}. Some recent methods query a large pre-trained model to directly evaluate alignment ~\cite{lin2024evaluating,chan2023clair,lee2024fleur,xu2024visionrewardfinegrainedmultidimensionalhuman,rambhatlaselfeval}. Although current metrics increasingly align with human preferences for visio-linguistic reasoning and text-to-image evaluation, many of these methods fail to evaluate longer, more descriptive captions effectively.
Most similar to our method, Image2Text2Image~\cite{huang2025image2text2image} computes image captioning performance by leveraging text-to-image generation to produce reconstructed images given text captions. The final score is the reconstruction error between the original and generated image's DINOv2~\cite{oquabdinov2} or CLIP~\cite{radford2021learning} features. DDPO~\cite{blacktraining} also includes a similar text-to-image-to text reconstruction score to optimize diffusion models. These pipelines match our dataset collection process outlined in Section~\ref{method}. However, our method uses cycle consistency scores to train a reward model with the benefit of inference speed, differentiability for downstream applications, and better performance.

\vspace{10pt}
\noindent
\textbf{Detailed captioning.}
Image-to-text models can produce comprehensive descriptions~\citep{team2023gemini,gpt-4o} by scaling the language model~\citep{liu2023llava,liu2024improved} and training on semantically rich synthetic captions~\citep{li2022blip,li2023blip,sharifzadeh2024synth,liu2023llava,liu2024improved}.
Despite growing model capabilities, little attention has been given to evaluating descriptive captions. Addressing this issue, DetailCaps-4870~\cite{dong2024benchmarking} evaluates image-text alignment metrics on detailed descriptions, whereas DeCapBench~\cite{yepainting} evaluates image-to-text models on detailed captioning using their reference-based metric DCSCORE. Our reward model provides a fast, differentiable, and reference-free approach to measuring alignment for descriptive texts.

\paragraph{Cycle consistency.}
Imposing cycle consistency continuously has been shown to be effective for many tasks in different domains~\cite{brislin1970back,godard2017unsupervised,he2016dual,huang2013consistent,wang2013image,yi2017dualgan,zach2010disambiguating,zhou2015flowweb,zhou2016learning,jabri2020space}, especially for self-supervised training and cases without paired ground truth annotations~\cite{godard2017unsupervised,hoffman2018cycada,li2023leveraging,messikommer2022bridging,yi2017dualgan,zhu2017unpaired,zach2010disambiguating,zhou2016learning,jabri2020space}, and recently for evaluating VLM and LLM performance~\cite{shah2019cycle,dhuliawala2024chain}. Rapid progression of multimodal models has facilitated exploring cycle consistency between images and texts~\cite{gorti2018text}, and incorporation of cycle consistency for training by combining text-to-image diffusion models and vision-language models~\cite{betker2023improving,esser2024scaling,sharifzadeh2024synth,li2023leveraging}.

\paragraph{Preference optimization.}
There are many techniques to align model outputs with human preferences~\cite{stiennon2020learning,ouyang2022training} at training~\cite{schulman2017proximal,rafailov2023direct,shao2024deepseekmath} or test time~\cite{stiennon2020learning,nakano2021webgpt,jinnai2024regularized}. These approaches have been applied mostly to large language models and recently to vision-language models~\cite{yu2024rlhf,sun2024aligning,yu2024rlaif} and diffusion models~\cite{wallace2024diffusion,blacktraining,prabhudesai2023aligning}. Text-to-image alignment metrics such as Human Preference Score (HPS)~\cite{wu2023human,wu2023humanv2}, PickScore~\cite{kirstain2023pick}, and ImageReward~\cite{xu2024imagereward} all collect human preferences to train a reward model. VLFeedback~\cite{silkie} substitutes human feedback by using foundation models (e.g., GPT-4V) to annotate preferences~\cite{yepainting,silkie,zhao2023beyond,yu2024rlaif} and applies Direct Preference Optimization (DPO)~\cite{rafailov2023direct} with their dataset. Our method collects preferences from a new signal: cycle consistency, which is cheaper and more easily scalable. We apply our dataset both to reward modeling and preference learning via DPO, exhibiting competitive performance with models trained on human labels.
\section{Method} \label{method}
\subsection{Cycle Consistency as Preferences} 

Our goal is to learn preferences for image-text alignment without relying on human annotations. Prior approaches often use humans~\cite{kirstain2023pick,wu2023human,xu2024imagereward} or GPT-4V~\cite{silkie} to rank the quality of generated captions or images. Instead, we propose to derive preferences from \textit{cycle consistency}. Given image-to-text mapping $F:X\rightarrow Y$, 
we measure how well text $F(x)$ aligns with image $x$ by measuring how well backward mapping $G:Y\rightarrow X$ can reconstruct $x$. We define \textit{cycle consistency score} for $F(x)$ conditioned on $x$ as:
\begin{equation}
    s(x \rightarrow F(x)) \defeq d_\text{img}(x,G(F(x))),
\end{equation}
where $d_{\text{img}}$ measures the similarity between the reconstructed image $G(F(x))$ and the original image $x$. We use DreamSim~\cite{fu2023dreamsim} to compute this similarity.

Similarly, for text-to-image mapping $G:Y\rightarrow X$, we measure how well image $G(y)$ aligns with text $y$ by using a backward mapping $F: X \rightarrow Y$. We define the cycle consistency score for $G(y)$ conditioned on $y$ as:
\begin{equation}
    s(y\rightarrow G(y)) \defeq d_\text{text}(y,F(G(y))),
\end{equation}
where $d_{\text{text}}$ measures the similarity between the reconstructed text $F(G(y))$ and the original text $y$. We use SBERT~\citep{reimers-2019-sentence-bert} to compute this similarity.

Importantly, these scores generalize to arbitrary image–text pairs $(x,y)$, not just model outputs:
\begin{equation} \label{eqn:backcasting1}
\begin{split}
    s(x \rightarrow y) &\defeq d_\text{img} (x, G(y)), \\
    s(y\rightarrow x) &\defeq d_\text{text}(y,F(x)).
\end{split}
\end{equation}
While prior work~\cite{garg2024imageinwords,huang2025image2text2image} uses this score directly as an alignment metric, we \textit{learn} alignment from a large pool of comparisons.
Given triplets $(x,y_i,y_j)$ and $(y,x_i,x_j)$, we convert cycle consistency scores into pairwise preferences:
\begin{equation} \label{eqn:pref1}
\begin{split}
y_i\succ y_j \;\; \text{if} \;\; s(x \rightarrow y_i) > s(x \rightarrow y_j), \\
x_i\succ x_j \;\; \text{if} \;\; s(y \rightarrow x_i) > s(y \rightarrow x_j).
\end{split}
\end{equation} 
where $\succ$ denotes that $y_i$ is preferred over $y_j$, vice versa. We establish the connection between cycle consistency score and cycle consistency of mappings in Appendix~\ref{appendix:cc_details}.

\begin{table}[t]
  \centering
  \resizebox{1.0\columnwidth}{!}{
  \begin{tabular}{lcccccc}
    \toprule
    Dataset & Task & \# Pairs & Supervision & Tokens \\ 
    \midrule
    ImageRewardDB~\cite{xu2024imagereward} & T2I & 137K & Human & 35.73 \\
    HPDv2~\cite{wu2023humanv2} & T2I & 798K & Human & 18.89 \\
    Pick-A-Pic v2~\cite{kirstain2023pick} & T2I & 851K & Human & 23.74 \\
    VLFeedback~\cite{silkie} & VL & 399K & GPT-4V~\cite{gpt-4v} & 97.03 \\
    \midrule
    CyclePrefDB-I2T & I2T & 398K & Cycle consistency & 56.82 \\
    CyclePrefDB-T2I & T2I & 468K & Cycle consistency & 55.13 \\
    \bottomrule
  \end{tabular}}
  \caption{\textbf{Key differences of preference datasets.} Existing preference datasets use human or GPT-4V annotations for supervision, whereas we label preferences with cycle consistency. We provide comparison pairs for both image-to-text (I2T) and text-to-image (T2I) tasks. CyclePrefDB features significantly denser text than typical T2I datasets, while remaining within token limits (77 tokens) of text-to-image models. VL denotes vision-language tasks.}
  \label{tab:data_comparison}
  \vspace{-0.4cm}
\end{table}

\subsection{Dataset Generation}

We design our dataset to capture alignment between images and \textit{dense} text, focusing on captioning images with rich descriptions and generating images from longer, detailed prompts. To this end, we use the train split of Densely Captioned Images (DCI) dataset~\citep{urbanek2024picture} for input images and texts. It contains 7.6K image-text pairs featuring high-resolution images annotated with dense captions. Due to prompt length constraints of text-to-image models, we use sDCI, a summarized version of DCI to fit within 77 tokens. See Appendix~\ref{appendix:recon_dataset} for details and visualizations.

\vspace{10pt}
\noindent
\textbf{Image-to-text generation.} Given image $x$, we first obtain multiple candidate text descriptions $\{y_1, ..., y_n\}$ of varying quality. In practice, we use 11 image-to-text models trained on different datasets and scales: BLIP2 (T5-XXL)~\cite{li2023blip}, LLaVA-1.5 (7B, 13B)~\cite{liu2024visual}, LLaVA-1.6 (7B, 34B)~\cite{liu2024improved}, LLaVA-OneVision (0.5B, 7B)~\cite{li2024llava}, and InternVL2 (2B, 8B, 26B, 40B)~\cite{internvl2,chen2024far}. As reward modeling is inherently contrastive, we deliberately include older models that produce short, hallucinated captions as negative examples alongside newer models to maximize text diversity. We specifically instruct the models to generate \textit{rich, descriptive} captions, using the prompt recommended by the model distributor (Appendix~\ref{appendix:prompts}). We use greedy sampling with a maximum token length of 77, i.e., maximum prompt length supported by the text-to-image models. We fix the backward mapping $G$ as Stable Diffusion 3 to compute $s(x\rightarrow y)$.

\vspace{10pt}
\noindent
\textbf{Text-to-image generation.} 
Given a text prompt $y$, we generate a set of image candidates $\{x_1, ..., x_n\}$ using 4 text-to-image models: Stable Diffusion 1.5~\citep{rombach2022high}, Stable Diffusion XL~\citep{podell2023sdxl}, Stable Diffusion 3~\citep{esser2024scaling}, and FLUX (Timestep-distilled)~\cite{FLUX22}. Similarly, we select models with varying performance to maximize diversity of generated images. We use three random seeds to generate the images, creating 12 candidate images per prompt. We fix the backward mapping $F$ as LLaVA-1.5-13B to compute $s(y\rightarrow x)$.

\subsection{Reward Modeling} \label{sec:reward_model}
The generality of cycle-consistent preferences allows us to train a reward model in multiple ways. We explore three variants: (1) \textbf{CycleReward-I2T}: trained with image-to-text preferences $s(x \rightarrow y)$, (2) \textbf{CycleReward-I2T}: trained with text-to-image preferences $s(y \rightarrow x)$, and (3) \textbf{CycleReward-Combo}: jointly trained on both datasets. 

\paragraph{Training details.}
Given a dataset of image-to-text comparisons $(x, y_i, y_j)$, where image $x$ is paired with preferred text $y_i$ and rejected text $y_j$, the loss is formulated as:
\begin{equation}
\mathcal{L}_\mathrm{img}\defeq - \mathbb{E}_{(x, y_i, y_j) \sim D_X} \left[ \log \sigma \left( r_{\theta}(x, y_i) - r_{\theta}(x, y_j) \right) \right],
\end{equation}
where $r_{\theta}(x, y)$ is the scalar output of the reward model~\cite{ouyang2022training,stiennon2020learning}. Similarly, given a dataset of text-to-image comparison pairs $(y, x_i, x_j)$, where text $y$ is paired with a preferred image $x_i$ and rejected image $x_j$, the loss is formulated as: 
\begin{equation}
\mathcal{L}_\mathrm{text} \defeq - \mathbb{E}_{(y, x_i, x_j) \sim D_Y} \left[ \log \sigma \left( r_{\theta}(x_i, y) - r_{\theta}(x_j, y) \right) \right].
\end{equation}
Finally, we also train a reward model on both datasets using the objective below. We set $\lambda=1$ for joint training.
\begin{equation}
\mathcal{L} = \mathcal{L}_\mathrm{text} + \lambda \mathcal{L}_\mathrm{img}.
\end{equation}

\paragraph{Network architecture.} Similar to ImageReward~\cite{xu2024imagereward}, we adopt BLIP~\cite{li2022blip} as our backbone. It consists of a ViT-L/16 encoder~\cite{dosovitskiy2020image} and a $\text{BERT}_\text{base}$ text encoder~\cite{devlin2019bert} followed by a 5-layer MLP. Training details are outlined in Appendix~\ref{appendix:training_details}.

\section{Cycle Consistency and Human Preferences}
Does cycle consistency align with human preferences? We measure the agreement rate between cycle consistency and human preferences on detailed captioning and text-to-image generation. For detailed captioning, we compare to human preferences from RLHF-V~\cite{yu2024rlhf} and POVID~\cite{zhou2024aligning} datasets.
For text-to-image generation, we compare to HPDv2~\cite{wu2023humanv2}, Pick-a-Pic v2~\cite{kirstain2023pick}, and ImageRewardDB~\cite{xu2024imagereward}. For each dataset, we sample 1K random binary comparison pairs. We compare human labels to raw cycle consistency scores, $s(x \rightarrow y)$ and $s(y \rightarrow x)$, as well as our trained reward models. We also compare against GPT-4o~\cite{gpt-4o} annotations, as they have been shown effective for preference learning~\cite{silkie}.

Table~\ref{tab:human_prefs}
shows that CycleReward achieves the highest agreement with human annotations, with CycleReward-Combo having the highest average agreement rate of 65\%.
While GPT-4o annotations on detailed captioning align more closely with humans, agreement drops significantly on text-to-image generation, with as low as 24.84\% on ImageRewardDB. In contrast, raw cycle consistency has a consistent agreement rate across both tasks. Training a reward model with cycle consistency further improves alignment, demonstrating the effectiveness of distilling cycle-consistent preferences into a learned reward model.

While we compare against human preferences, our aim is not to mimic them. Instead, we aim to learn \textit{image-text alignment}, and demonstrate that cycle consistency is an effective proxy---achieving strong results without collecting \textit{any} human labels, as shown in the following sections.

\begin{table}[t]
  \centering
  \resizebox{1.0\columnwidth}{!}{
  \begin{tabular}{l|cc|ccc}
    \toprule
      & \multicolumn{2}{c|}{\textit{Detailed Captioning}} & \multicolumn{3}{c}{\textit{Text-to-Image Generation}} \\
    Method & RLHF-V & POVID & HPDv2 & PaPv2 & IRDB \\
    \midrule
    GPT-4o & 61.3 & 60.0 & 48.1 & 45.8  & 24.8 \\
    Raw Cycle Consistency & 58.6 & 61.2 & 60.5 & 59.8  & 54.5 \\
    \midrule
    \cellcolor{myapricot}CycleReward-I2T & \cellcolor{myapricot}63.9 & \cellcolor{myapricot}65.6 & \cellcolor{myapricot}66.5 & \cellcolor{myapricot}65.7 & \cellcolor{myapricot}60.2 \\
    \cellcolor{myapricot}CycleReward-T2I & \cellcolor{myapricot}57.1 & \cellcolor{myapricot}\textbf{78.2} & \cellcolor{myapricot}\textbf{68.3} & \cellcolor{myapricot}\textbf{66.2} &	\cellcolor{myapricot}60.2 \\
    \cellcolor{myapricot}CycleReward-Combo & \cellcolor{myapricot}\textbf{66.5} &	\cellcolor{myapricot}63.8 & \cellcolor{myapricot}67.7 &	\cellcolor{myapricot}65.8 &	\cellcolor{myapricot}\textbf{61.3} \\
    \bottomrule
  \end{tabular}}
  \vspace{-0.2cm}
  \caption{Agreement rates (\%) between human preferences and those from GPT-4o, raw cycle consistency, and CycleReward.}
  \vspace{-0.4cm}
  \label{tab:human_prefs}
\end{table}

\section{Reward Model Evaluation} \label{sec:benchmarks}
We evaluate CycleReward's ability to \textit{assess} and \textit{improve} image-text alignment across two tasks: detailed captioning and text-to-image generation.
Specifically, we evaluate CycleReward as an alignment metric,
and then deploy it to maximize inference-time alignment via Best-of-$N$.

\paragraph{Comparison methods.} \label{comparison_methods}
We compare against current reference-free image-text alignment metrics. These include: (1) \textbf{CLIPScore}~\cite{hessel2021clipscore} which measures cosine similarity between image and text embeddings from CLIP~\cite{radford2021learning}, (2) \textbf{ImageReward}~\cite{xu2024imagereward}, 
(3) \textbf{HPSv2}~\cite{wu2023human,wu2023humanv2} and (4) \textbf{PickScore}~\cite{kirstain2023pick}, which are trained on large human preference datasets for text-to-image generation, and
(5) \textbf{VQAScore}~\cite{lin2024evaluating} which produces alignment scores by querying a VLM with the prompt \texttt{"Does this figure show \{text\}?"}. For VQAScore, we compare two different model sizes: CLIP-T5-xl (3B) and CLIP-T5-xxl (11B). (6) \textbf{Raw cycle consistency} directly uses alignment scores $s(x \rightarrow y)$ for image-to-text generation and $s(y \rightarrow x)$ for text-to-image generation without learning a reward model. For image-to-text, this is equivalent to Image2Text2Image~\cite{huang2025image2text2image}. We adopt the same model configurations (i.e., decoders, similarity metrics) for fair comparison.

\begin{table}[t]
  \centering
  \resizebox{1.0\columnwidth}{!}{
  \begin{tabular}{l|cc}
    \toprule
    Method & DetailCaps-4870 & GenAI-Bench \\
    \midrule
    \textit{Vision-Language Model}\\
    CLIPScore & 51.66 & 49.73\\
    VQAScore (3B) & 46.84 & 59.54\\
    VQAScore (11B) & 50.24 & \textbf{64.13} \\
    \midrule
    \textit{Human Preferences} \\
    HPSv2 & 54.34 & 56.13 \\
    PickScore & 51.01 & 57.05 \\
    ImageReward & 50.70 & 56.70 \\
    \midrule
    \textit{Cycle Consistency} \\
    Raw Cycle Consistency & 56.46 & 52.52 \\
    IRDB-Cycle & 49.96 & 54.58 \\
    \cellcolor{myapricot}CycleReward-I2T & \cellcolor{myapricot}58.02 & \cellcolor{myapricot}53.49 \\
    \cellcolor{myapricot}CycleReward-T2I & \cellcolor{myapricot} 51.74 & \cellcolor{myapricot}55.20 \\
    \cellcolor{myapricot}CycleReward-Combo & \cellcolor{myapricot}\textbf{60.50} & \cellcolor{myapricot}55.52 \\
    \bottomrule
  \end{tabular}}
  \vspace{-0.2cm}
  \caption{\textbf{Evaluating image-text alignment.} CycleReward-Combo and CycleReward-I2T outperform all approaches on detailed captioning evaluation, even those trained on human preferences. Notably, we outperform VQAScore with 24$\times$ larger model size. For text-to-image generation, CycleReward achieves similar performance to models trained on human preferences, while VQAScore outperforms others. Across both tasks, our \textit{learned} reward model outperforms using raw cycle consistency.}
  \vspace{-0.4cm}
  \label{tab:alignment_eval}
\end{table}

\begin{figure*}[t]
  \centering
  \includegraphics[width=1\linewidth]{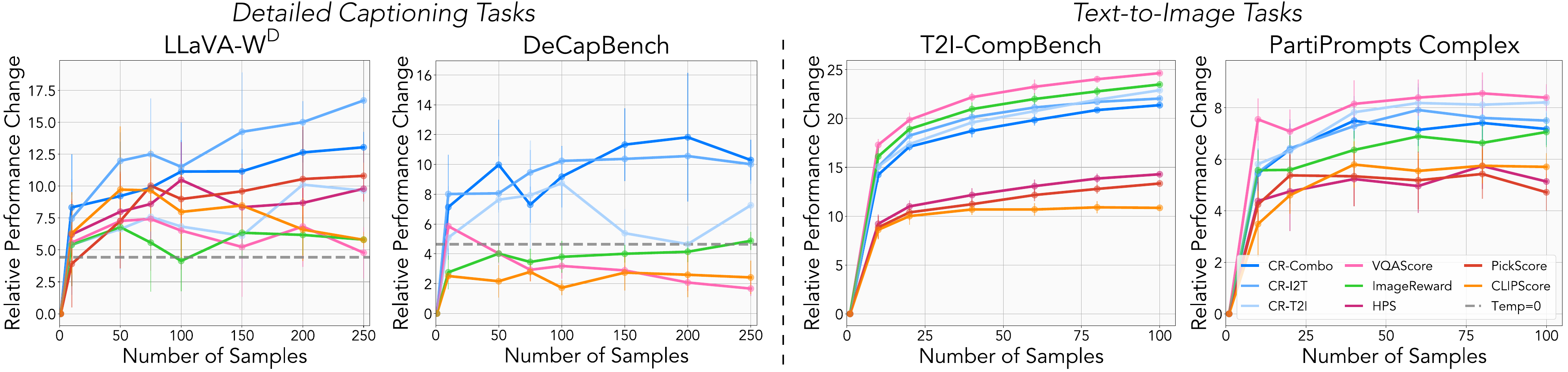}
  \vspace{-0.5cm}
  \caption{\textbf{Best-of-$N$ relative performance gain.} From left to right: LLaVA-W, DeCapBench, T2I-CompBench (mean of 6 categories), and PartiPrompts (complex). In each plot, we show the relative performance gain from BoN sampling with different metrics. Feedback from our reward model leads to the greatest overall improvement for detailed captioning tasks, while we maintain competitive text-to-image generation performance with VQAScore and ImageReward.}
  \vspace{-0.4cm}
  \label{fig:BoN_i2t}
\end{figure*}

\subsection{Metric for Image-Text Alignment} \label{sec:alignment}

\paragraph{Evaluation benchmarks.}
While many benchmarks exist for short captions, few target detailed descriptions, and those that do often lack labels or contain limited examples. One exception is DetailCaps-4870~\cite{dong2024benchmarking}, which evaluates captions on accuracy and inclusion across object, attribute, and relation categories. It contains 4,870 image-text pairs from ShareGPT4V~\cite{chen2024sharegpt4v}, LLaVA 1.5, and CogVLM~\cite{hong2023cogagent,wang2023cogvlm}, scored by three VLMs: GPT-4V~\cite{gpt-4v}, Gemini-1.5 Pro~\cite{team2024gemini}, and GPT-4o~\cite{gpt-4o}. We use the mean score as a pseudo-ground truth. To evaluate text-to-image generation, we use GenAI-Bench~\cite{lin2024evaluating,li2024genaibench}, which consists of 1,600 prompts paired with 6 generated images from different models. Each generation is annotated with three human ratings based on fidelity to the text. For both tasks, we measure agreement with alignment metrics using pairwise accuracy~\cite{deutsch2023ties}.

\paragraph{Comparison to human preference learning.} 
We directly compare to ImageReward, which uses human labels, by training a reward model on the \textit{same} backbone and ImageRewardDB dataset, but re-annotated with cycle consistency preferences. We refer to this model as IRDB-Cycle. Table~\ref{tab:alignment_eval} shows IRDB-Cycle achieves comparable performance to ImageReward, demonstrating that cycle consistency is an effective and cheaply scalable alternative for human labels. In the following sections, we show training on \textit{our dataset} yields further improvements.

\paragraph{Results.} 
Table~\ref{tab:alignment_eval} reports pairwise accuracy between different methods and human preferences. For detailed captioning, CycleReward outperforms all existing methods by a large margin, including HPSv2, PickScore, and ImageReward, which are trained on \textit{human} preferences. Notably, CycleReward outperforms VQAScore (11B) by 10.26\%, which is a 24$\times$ larger model. It outperforms raw cycle consistency,
which highlights the effectiveness of distilling cycle consistency into a learned reward model.

For text-to-image generation, CycleReward performs comparably to HPS, PickScore, and ImageReward, all of which are trained with human annotations. CycleReward outperforms both raw cycle consistency and IRDB-Cycle, a model trained on ImageRewardDB with cycle-consistent labels. Although VQAScore (11B) aligns most with humans, our model does surprisingly well considering its small scale (477M). See Appendix~\ref{appendix:benchmark_examples} for qualitative comparisons. 

\subsection{Best-of-N Sampling} 
\label{sec:boN}
Best-of-$N$ (BoN) sampling is a simple strategy to improve model results at test time~\cite{cobbe2021training,stiennon2020learning,ma2025inference}. The process involves generating $N$ candidate outputs from a base model, ranking them using a reward model, and selecting the one with the highest score. The selection criterion is entirely based on the reward model, and naturally better models choose higher-quality outputs.

\paragraph{Evaluation benchmarks.}
For image-to-text generation, we use two detailed captioning benchmarks: LLaVA-W~\cite{liu2023llava} detailed captioning subset (LLaVA-W\textsuperscript{D}) and DeCapBench~\cite{yepainting}, which assess the correctness and coverage of details (i.e., precision and recall) in generated captions. 
LLaVA-W\textsuperscript{D} evaluations are conducted using GPT-4o-mini~\cite{gpt-4o} as the evaluator model, while DeCapBench uses DCScore~\cite{yepainting}. For text-to-image generation, we use T2I-Compbench~\cite{huang2023t2i} for fine-grained preferences on six compositional categories, and the ``complex'' subset of PartiPrompts~\cite{yuscaling} for complex, detailed prompts.

\paragraph{Detailed captioning results.} \label{BoN}
For each image, we perform BoN selection from a pool of 250 captions obtained from a combination of temperature, nucleus, and prompt sampling LLaVA1.5-13B~\cite{li2024llava,liu2024visual} (see Appendix~\ref{appendix:bon_cont} for details). For image captioning, BoN sampling with our reward model increases performance significantly over other metrics as seen in Figure~\ref{fig:BoN_i2t}. Both LLaVA-W\textsuperscript{D} and DeCapBench assess captions based on correctness and level of detail, and our reward model yields the largest improvement in the overall evaluation score. In Appendix~\ref{appendix:bon_cont} we plot BoN results for the non-hallucination and comprehensiveness scores from DeCapBench and find that our model excels at describing many things in detail while maintaining correctness (albeit less accurately than VQAScore). In contrast, baselines such as VQAScore and ImageReward highly weigh accuracy to the point of preferring captions with significantly less detail.

\vspace{10pt}
\noindent
\textbf{Text-to-image generation results.}\label{sec:t2i_bon}
For all text prompts, we use SDXL-Turbo~\cite{sauer2023adversarial} to generate a pool of 100 images with different random seeds to perform BoN sampling. Figure~\ref{fig:BoN_i2t} (right) shows relative performance gain using different reward models for BoN sampling. Note that our self-supervised reward models perform similarly to ImageReward which is trained with human preferences, and even outperforms on ``complex'' text prompts. 
For specific T2I-CompBench category results see Appendix~\ref{appendix:t2icb}.

\begin{figure*}[t]
  \centering
  \includegraphics[width=0.99\linewidth]{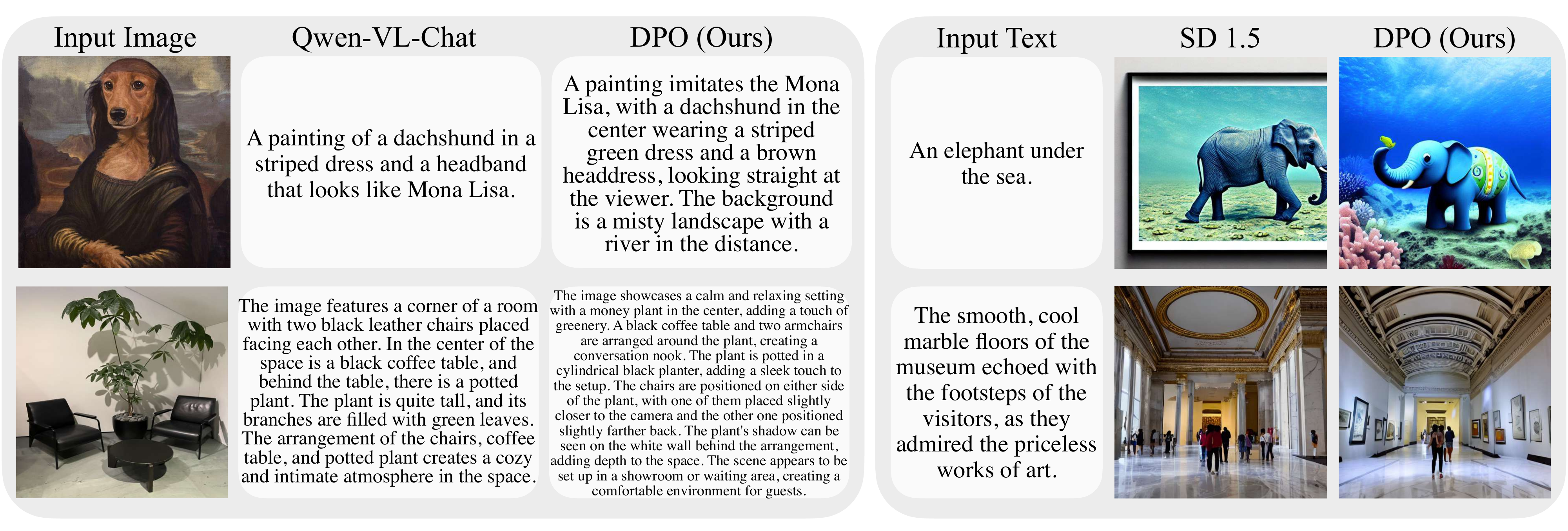}
  \vspace{-0.2cm}
  \caption{\textbf{DPO results using CyclePrefDB.} (Left) Using CyclePrefDB-I2T for DPO improves Qwen-VL-Chat, yielding denser captions that describe fine-grained details of the input image. (Right) Using CyclePrefDB-T2I for Diffusion DPO improves Stable Diffusion 1.5, producing images that better capture the details in the input prompt.} 
  \vspace{-0.2cm}
  \label{fig:detailcapsbench}
\end{figure*}

\subsection{Ablation Study} \label{ablations}
We ablate several design choices for our reward model on DetailCaps-4870 and GenAI-Bench. For both benchmarks, we report pairwise accuracy, and \colorbox{mygray}{gray} entries denote design choices used in CycleReward. For ablations on objective function, data scale, and filtering see Appendix~\ref{appendix:more_ablations}.

\begin{table}[h]
  \centering
  \resizebox{1.0\columnwidth}{!}{
  \begin{tabular}{lcc}
    \toprule
    Metric & DetailCaps-4870 & GenAI-Bench \\
    \midrule
    \textit{Image similarity metric} \\
    \cellcolor{mygray}{DreamSim} & \cellcolor{mygray}\textbf{58.02} & \cellcolor{mygray}\textbf{53.49}\\
    LPIPS & 53.16 & 52.97\\
    CLIP & 57.90 & 53.30\\
    \midrule
    \textit{Text similarity metric} \\
    \cellcolor{mygray}{SBERT} & \cellcolor{mygray}\textbf{51.74} & \cellcolor{mygray}55.20\\
    BERT & 47.27 & \textbf{55.52}\\
    CLIP & 49.00 & 54.92\\
    \bottomrule
  \end{tabular}}
  \vspace{-0.2cm}
  \caption{\textbf{Effect of similarity metrics} for comparing original inputs and reconstructions. Choices used by our model are in \colorbox{mygray}{gray}.}
  \vspace{-0.6cm}
  \label{tab:ablation_metric}
\end{table}

\paragraph{Similarity metric.} We study the effect of different image and text similarity metrics for computing cycle consistency scores $s(x \rightarrow y)$ and $s(y \rightarrow x)$. For image similarity, we compare DreamSim, LPIPS~\cite{zhang2018unreasonable}, and CLIP~\cite{radford2021learning}, where we compute cosine similarity between CLIP image embeddings. For text similarity, we compare SBERT with BERTScore~\cite{zhang2019bertscore} and CLIP~\cite{radford2021learning} text embedding cosine similarity. The ablation study justifies our choices, as DreamSim and SBERT achieve the best average performance across image-to-text and text-to-image tasks. In particular, DreamSim models human visual similarity which may contribute to better alignment judgments.

\vspace{10pt}
\noindent
\textbf{Decoders.} We examine the effect of decoders for generating image and text reconstructions. For text-to-image decoders, we compare Stable Diffusion 3, Flux-Schnell~\cite{FLUX22} and SDXL-Turbo~\cite{podell2023sdxl,rombach2022high}. We find that Stable Diffusion 3, with more denoising steps, achieves best performance for detailed captioning, while SDXL-Turbo has a slight edge on text-to-image generation. For image-to-text decoders, we compare LLaVA-1.5 13B, LLaVA-OV-7B~\cite{li2024llava}, InternVL-26B~\cite{internvl2}. Using InternVL2-26B, with its larger, more performant language model, significantly improves detailed captioning evaluation, with similar performance in text-to-image generation. These results suggest that improvements in decoder quality can further enhance the effectiveness of cycle consistency as a supervised signal for alignment.

\begin{table}[h]
  \centering
  \resizebox{1.0\columnwidth}{!}{
  \begin{tabular}{lcc}
    \toprule
    Decoder & DetailCaps-4870 & GenAI-Bench \\
    \midrule
    \textit{Text-to-image decoder}\\
    \cellcolor{mygray}{Stable Diffusion3} & \cellcolor{mygray}\textbf{58.02} & \cellcolor{mygray}53.49\\
    FluxSchnell & 56.54 & 53.19\\
    SDXL-Turbo & 56.42 & \textbf{54.83} \\
    \midrule
    \textit{Image-to-text decoder}\\
    \cellcolor{mygray}{LLaVA-1.5-13B} & \cellcolor{mygray}51.74 & \cellcolor{mygray}\textbf{55.20}\\
    LLaVA-OV-7B & 52.80 & 53.09\\
    InternVL2-26B & \textbf{57.21} & 54.46\\
    \bottomrule
  \end{tabular}}
  \vspace{-0.2cm}
  \caption{\textbf{Effect of decoder models} for generating reconstructions. Choices used by our model are in \colorbox{mygray}{gray}.}
  \vspace{-0.3cm}
  \label{tab:ablation_decoder}
\end{table}


\begin{table*}[t]
  \centering
  \resizebox{1.0\textwidth}{!}{
  \begin{tabular}{lcc|cccccc}
    \toprule
      & \multicolumn{2}{c|}{\textit{Detailed Captioning}} & \multicolumn{5}{c}{\textit{General VQA Tasks}} \\
    Model & DeCapBench & LLaVA-W\textsuperscript{D} & LLaVA-W\textsuperscript{C} & LLaVA-W\textsuperscript{R} & MMHalBench & MME\textsuperscript{P} & MME\textsuperscript{C} \\
    \midrule
    Qwen-VL-Chat & 26.47 & 61.67 & 73.10 & 83.71 & 2.99 & 1460.2 & 368.9\\
    DPO w/ VLFeedback & 28.03 &  69.17 &\textbf{76.39} & \textbf{89.50} & \textbf{3.32} & \textbf{1551.5} & \textbf{396.8} \\
    \midrule
    \cellcolor{myapricot}DPO w/ CyclePrefDB-I2T & \cellcolor{myapricot}\textbf{30.63} &  \cellcolor{myapricot}\textbf{70.00} &\cellcolor{myapricot}74.13 & \cellcolor{myapricot}84.62 & \cellcolor{myapricot}3.11 & \cellcolor{myapricot}1485.7 & \cellcolor{myapricot}386.4 \\
    \bottomrule
  \end{tabular}}
  \caption{\textbf{Direct preference optimization (DPO) for image-to-text generation.} The best results are indicated in \textbf{bold}. DPO with CyclePrefDB-I2T improves the base model's performance across all tasks---including detailed captioning, perception, reasoning, and hallucination reduction---despite only containing captioning instructions. It achieves comparable or superior results to VLFeedback, a preference dataset annotated with GPT-4V spanning diverse task instructions.}
  \label{tab:i2t_dpo}
\end{table*}

\begin{table*}[t]
  \centering
  \resizebox{1.0\textwidth}{!}{
  \begin{tabular}{lcccccc|cc|cc}
    \toprule
     & \multicolumn{6}{c|}{\textit{T2I-CompBench}} & \multicolumn{2}{c|}{\textit{Short Prompts}} & \multicolumn{2}{c}{\textit{Long Prompts}}\\
    Model & Spatial & Color & Complex& Numeracy & Shape & Texture & DrawBench & PP-Simple Detail & PP-FG Detail & PP-Complex \\
    \midrule
    Stable Diffusion 1.5 & 11.49 & 36.98 & 34.49 & 44.81 & 37.48 & 40.39 & 28.42 & 7.65 & 7.13 & 6.37 \\
    Diffusion DPO w/ Pick-A-Pic & 14.59 & 39.12 & 34.69 & \textbf{45.88} & 37.39 & 40.66 & \textbf{30.13} & \textbf{7.73} & \textbf{7.28} & 6.45\\
    \midrule
    \cellcolor{myapricot}Diffusion DPO w/ CyclePrefDB-T2I & \cellcolor{myapricot}\textbf{16.55} & \cellcolor{myapricot}\textbf{42.35} & \cellcolor{myapricot}\textbf{37.75} & \cellcolor{myapricot}45.24 & \cellcolor{myapricot}\textbf{38.83} & \cellcolor{myapricot}\textbf{46.67} & \cellcolor{myapricot}30.04 & \cellcolor{myapricot}7.69 & \cellcolor{myapricot}7.28 & \cellcolor{myapricot}\textbf{6.51} \\
    \bottomrule
  \end{tabular}}
  \caption{\textbf{Direct preference optimization (DPO) for text-to-image generation.} For all evaluations, higher scores are better. T2I-Compbench and DrawBench scores range from 0 to 100 while PartiPrompt (PP) scores range from 1 to 10. In all cases, the Diffusion DPO training with CyclePrefDB-T2I outperforms the base model. Furthermore, our model often outperforms or is comparable with the Pick-A-Pic Diffusion DPO model, especially for longer text prompts.}
  \label{tab:t2i_dpo}
  \vspace{-0.4cm}
\end{table*}

\section{Direct Preference Optimization} \label{sec:dpo}
We study the alignment effect of cycle-consistent preferences with direct preference optimization (DPO)~\cite{rafailov2023direct}, which optimizes the model to
prefer the chosen response over the rejected one without explicit reward modeling.
For image-to-text generation, we apply DPO~\cite{rafailov2023direct} to Qwen-VL-Chat~\cite{bai2023qwen} using CyclePrefDB-I2T. For text-to-image generation, we apply Diffusion DPO~\cite{wallace2024diffusion} to Stable Diffusion 1.5~\cite{rombach2022high} using our CyclePrefDB-T2I dataset. For implementation details see Appendix~\ref{appendix:training_details}.

\vspace{10pt}
\noindent
\textbf{Comparison methods.} We compare against the base model and models trained on different preference datasets. For image-to-text generation, we compare against VLFeedback~\cite{silkie}, a vision-language feedback dataset annotated with GPT-4V. It comprises 82K instructions, including visual question answering, image captioning and classification, reasoning, conversation, and red teaming, totaling 399K preference pairs. For text-to-image generation, we compare against Pick-A-Pic v2~\cite{kirstain2023pick}, a human preference dataset for text-to-image generation comprising 851K comparison pairs for 58,960 unique text prompts. 
Note that both datasets are larger than CyclePrefDB, which consists of 398K image-to-text pairs and 468K text-to-image pairs.

\vspace{10pt}
\noindent
\textbf{Evaluation benchmarks.} 
We evaluate on LLaVA-W\textsuperscript{D}~\cite{liu2023llava} and DeCapBench~\cite{yepainting} for detailed captioning. Although our dataset focuses on detailed captioning, we test generalization to new tasks: MME~\cite{fu2023mme} consists of MME\textsuperscript{P} for perception abilities and MME\textsuperscript{C} for cognition abilities such as coding and math problems, MMHal-Bench~\cite{sun2024aligning} for hallucination, and LLaVA-W\textsuperscript{C} for conversation capabilities and LLaVA-W\textsuperscript{R} for reasoning. For text-to-image generation, 
we use T2I-Compbench~\cite{huang2023t2i} for compositionality, 
DrawBench~\cite{saharia2022photorealistic} for general short prompts, and PartiPrompts~\cite{yuscaling} for dense prompts using the ``simple detail,'' ``fine-grained detail'' and ``complex'' categories.
For each prompt, we generate 10 images from different random seeds. To reduce variance, we repeat DrawBench and PartiPrompts GPT-4o evaluations 
five times and report mean scores.

\subsection{Results}
\paragraph{Image-to-text generation.}
To our surprise,
DPO fine-tuning with CyclePrefDB-I2T enhances the base model's performance across \textit{all} vision-language tasks---including detailed captioning, perception, reasoning, and hallucination---although our dataset only contains captioning instructions. Despite our narrow task instruction and smaller dataset size, it achieves comparable or superior results to VLFeedback, a preference dataset annotated by GPT-4V across VQA, captioning, classification, reasoning, conversation, and red teaming instructions.

\paragraph{Text-to-image generation.}
Table~\ref{tab:t2i_dpo} reports evaluation results on T2I-CompBench and DrawBench (scores from 1 to 100) and PartiPrompts (scores from 1 to 10), where higher is better. Across all categories, the model trained on CyclePrefDB-T2I outperforms the base model and is comparable with or outperforms the Pick-A-Pic model especially on complex prompts, which is particularly a challenge for Stable Diffusion 1.5. See Figure~\ref{fig:detailcapsbench} and Appendix~\ref{fig:ddpo_t2i} for qualitative results.


\section{Discussion}
We find that cycle consistency provides a scalable and effective supervisory signal for image-text alignment, achieving competitive performance without relying on any human-labeled data. We first construct CyclePrefDB, a preference dataset annotated via cycle consistency, and then train reward models that generalize across both image-to-text and text-to-image tasks. These models outperform or match existing baselines on detailed captioning and compositional text-to-image benchmarks, suggesting that cycle consistency is an effective alternative to human annotations. 

However, our method has limitations. Supervision quality depends on accurate reconstructions from pre-trained decoders, and generation errors can mislead preferences. Appendix~\ref{appendix:failures} visualizes failures cases and discusses more limitations. Future work could address these challenges by improving reconstructions, prompt diversity, and applying cycle consistency in different scenarios. Broadly, our framework offers a general approach for learning dense alignment between modalities, and could be extended to new domains such as audio-text, video-language, or even reasoning tasks.


\subsubsection*{Acknowledgments}
Research was sponsored by the Department of the Air Force Artificial Intelligence Accelerator and was accomplished under Cooperative Agreement Number FA8750-19-2-1000. The views and conclusions contained in this document are those of the authors and should not be interpreted as representing the official policies, either expressed or implied, of the Department of the Air Force or the U.S. Government. The U.S. Government is authorized to reproduce and distribute reprints for Government purposes notwithstanding any copyright notation herein. This work was supported in part by a Packard Fellowship and a Sloan Research Fellowship to P.I., by the MIT-IBM Watson AI Lab, by the Sagol Weizmann-MIT Bridge Program, by ONR MURI grant N00014-22-1-2740, by the MIT-Google program for computing innovation, the Amazon Science Hub, and an MIT-GIST grant.

{
    \small
    \bibliographystyle{ieeenat_fullname}
    \bibliography{main}
}

\clearpage
\appendix
\section*{Appendix}
\section{Cycle Consistency and Point-wise Mutual Information} \label{appendix:cc_details}

Let $X$ and $Y$ be random variables that take on realizations $x$ and $y$, respectively. In Section~\ref{method} $X$ and $Y$ represent images and texts, but note how our cycle consistency score (Equation~\ref{eqn:backcasting1}) and preference creation (Equations~\ref{eqn:pref1}) are general to any $X$ and $Y$. We now focus on the general case.

In Equation~\ref{eqn:backcasting1}, we define $s(x \rightarrow y)$ and $s(y \rightarrow x)$ with respect to fixed backward mappings $G:Y\rightarrow X$ and $F:X\rightarrow Y$ respectively. If $F,G$ are stochastic mappings, then we can view $G$ as sampling some $x'=G(y)$ from the distribution $p_G(X|Y=y)$ - a distribution which is determined by $G$. Symmetrically, we can view $F$ as sampling $y'=F(x)$ from the distribution $p_F(Y|X=x)$ determined by $F$. We then argue that distributionally,
\begin{equation}
    \begin{split}
        s(x \rightarrow y)_d &\defeq \log p_G(x|y)\\ 
        s(y \rightarrow x)_d &\defeq \log p_F(y|x) 
    \end{split}
\end{equation}
If the two distributions $p_F$ and $p_G$ sample from the same underlying distribution $p$, we can define joint distributional cycle consistency score. This may be the case if $F$ and $G$ are trained on the same dataset or with sufficient examples to model the same distributions.
\begin{equation}
\begin{split}
    s&(x, y)_d \defeq s(x \rightarrow y)_d + s(y \rightarrow x)_d \\
    &= \log p(x|y) + \log p(y|x) \;\;\;\;x,y\sim p(X,Y)
\end{split}
\end{equation}

\paragraph{Mutual Information}
Following the connection that previous work~\cite{li2017alice} has made between cycle consistency and mutual information, we rewrite the joint reward as follows:
\begin{equation}
\begin{split}
    s&(x, y)_d = \log p(x|y) + \log p(y|x) \\
    &= \log \frac{p(x,y)}{p(y)} + \log \frac{p(x,y)}{p(x)} \\
    &= \log \frac{p(x,y)^2}{p(x)p(y)} \\
    &= \log p(x,y) + \text{PMI}(x,y) \\
\end{split}
\end{equation}
Therefore, we can view the joint cycle consistency score as measuring both the likelihood of the pairing $p(x,y)$ and the pointwise mutual information. In turn, CycleReward prefers $x,y$ pairings which are both high probability and informative of each other.

\section{Benefits from Reward Modeling}
Because our reward model is trained with preferences from cycle consistency, it is natural to assume that the performance of raw cycle consistency scores $s(x\rightarrow y)$ and $s(y \rightarrow x)$ would be an upper bound for our reward model. In contrast, our trained reward models outperform raw cycle consistency on all benchmarks reported in Section~\ref{sec:benchmarks} in both mapping directions.
\begin{figure}[t]
  \centering
  \includegraphics[width=0.85\linewidth]{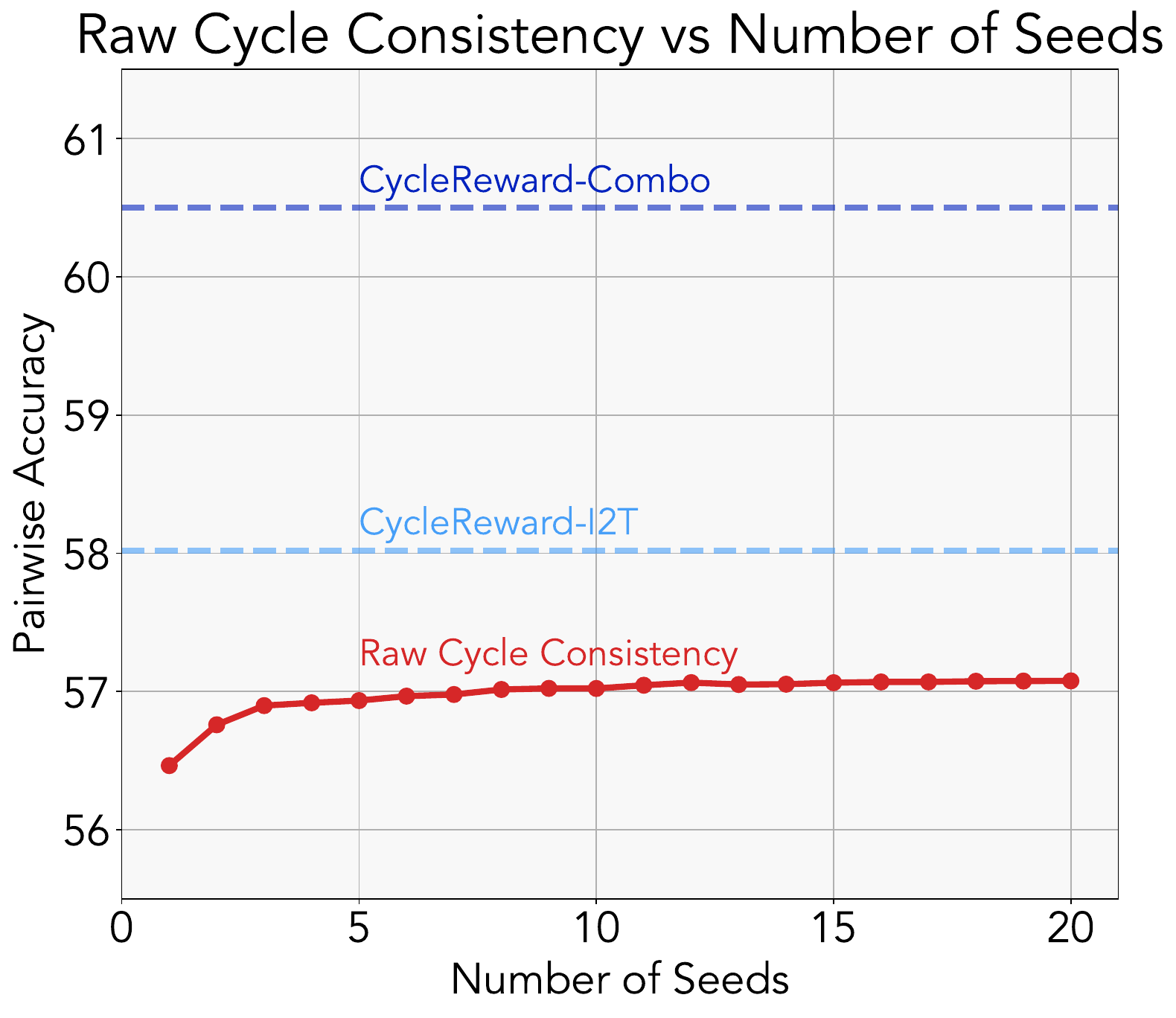}
  \vspace{-0.3cm}
  \caption{\textbf{Raw cycle consistency performance with increasing number of samples.} We plot DetailCaps-4870 benchmark performance (Pairwise Accuracy) for raw cycle consistency calculated over multiple samples (random seed sampling). Despite the increasing number of seeds, raw cycle consistency performance does not come close to reward model performance.} 
  \vspace{-0.4cm}
  \label{fig:rcc_vs_reward}
\end{figure}

\begin{figure*}[t]
  \centering
  \includegraphics[width=\linewidth]{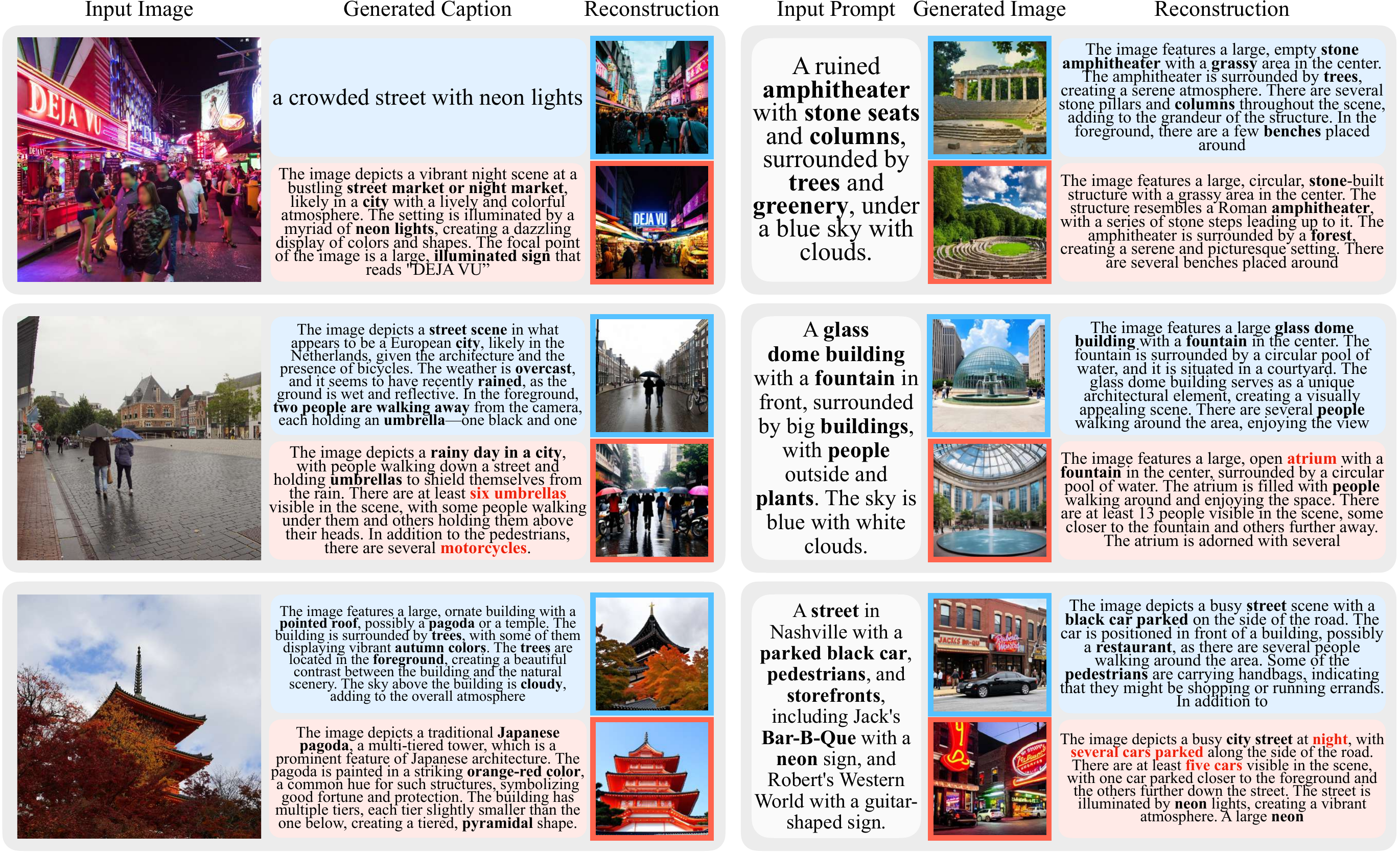}
  \caption{\textbf{Examples of CyclePrefDB}. \textcolor{cyan}{Preferred} samples are in blue and \textcolor{red}{rejected} samples are in red. (Left) We show input images, generated captions, and image reconstructions for image-to-text comparison pairs. (Right) shows input prompts, generated images, and text reconstructions for text-to-image comparison pairs. Generally, more accurate, descriptive captions and images that faithfully capture the prompt yield better reconstructions. However, exceptions exist such as the neon sign example (top left).}
  \vspace{-0.3cm}
  \label{fig:reconstructions}
\end{figure*}

Albeit computationally slow, averaging raw cycle consistency scores over multiple reconstructions as in Equation~\ref{eqn:mean_cc} could provide more accurate alignment measurements than just a single forward pass. We define the mean image-to-text cycle consistency as follows:
\begin{equation} \label{eqn:mean_cc}
    s^{*}(x \rightarrow y) = \frac{1}{N}\sum_{n=1}^{N}||x - g(y, z_n)||\;\;\;\;\;z_n \sim \mathcal{N}(0,I)
\end{equation}
This measurement averages $s(x \rightarrow y)$ scores over $N$ decoder reconstructions. In practice, we sample reconstructions by using different random seeds for the SD3 decoder. Note we can define a symmetric mean cycle consistency score for $s(y \rightarrow x)$, but focus on the image-to-text direction in this section.

Figure~\ref{fig:rcc_vs_reward} plots DetailCaps-4870 benchmark performance against the number of samples $N$ used to compute the mean cycle consistency score. Although using more seeds benefits raw cycle consistency, improvement tapers off around $N=5$ and never reaches the performance of CycleReward.

\begin{figure}[h]
  \centering
  \includegraphics[width=0.97\linewidth]{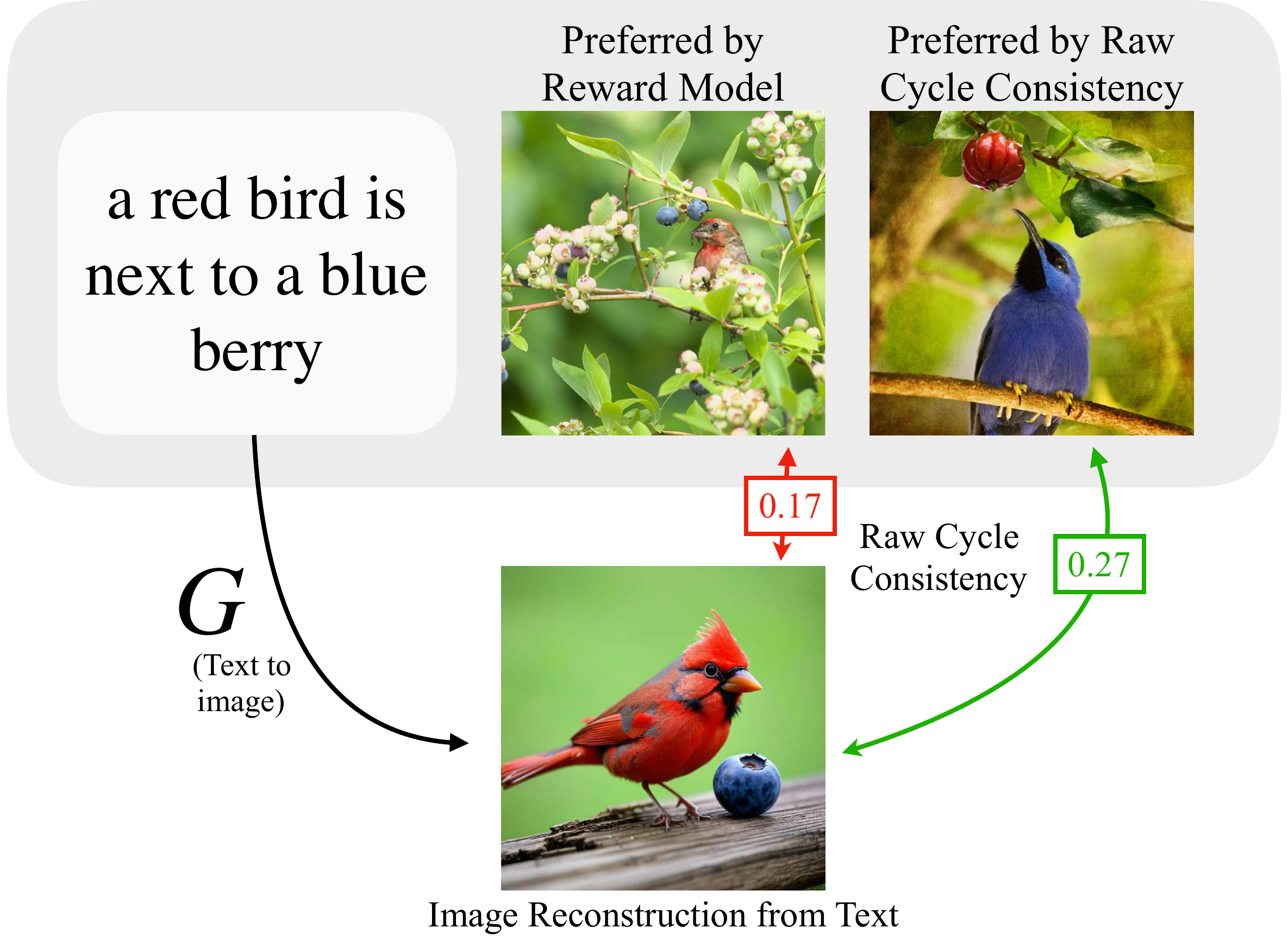}
  \vspace{-0.3cm}
  \caption{Raw cycle consistency $s(x\rightarrow y)$ is computed by comparing the original image (top) with its reconstruction (bottom), with similarity values shown in each box. In this example, although the reconstructed image accurately reflects the prompt, it is visually more similar to the image of the blue bird, leading to an incorrect alignment judgment based on raw cycle consistency. In contrast, our learned reward model, CycleReward, correctly identifies the true alignment.}  
  \vspace{-0.3cm}
  \label{fig:birb}
\end{figure}

Figure~\ref{fig:birb} qualitatively compares alignment computed by raw cycle consistency against our reward model. From the rich visual descriptions in our dataset, the reward model has learned that the image of the red bird corresponds best with the text description. In contrast, raw cycle consistency attempts to reconstruct the original input from the input prompt. Due to the lack of fine-grained visual information in the text, the reconstruction is more of a typical, object-centered bird image that happens to be structurally similar to the image of the blue bird over the red bird. This finding highlights additional benefits of distilling cycle consistency to a reward model -- beyond speed and differentiability. 

\section{CyclePrefDB Dataset Details} \label{appendix:recon_dataset}

\paragraph{Image and Text Reconstructions} \label{appendix:dataset_recons}
We provide examples of reconstructed images and texts used to create comparison pairs in our dataset in Figure~\ref{fig:reconstructions}. Generally, we find that better, more descriptive image captions lead to image reconstructions that are more similar to the input image. Symmetrically, generated images that are faithful to the prompt have text reconstructions reflecting this. However, failure cases can occur due to poor reconstructions as in Figure~\ref{fig:failures}.

\begin{figure}[t]
  \centering
  \includegraphics[width=0.98\linewidth]{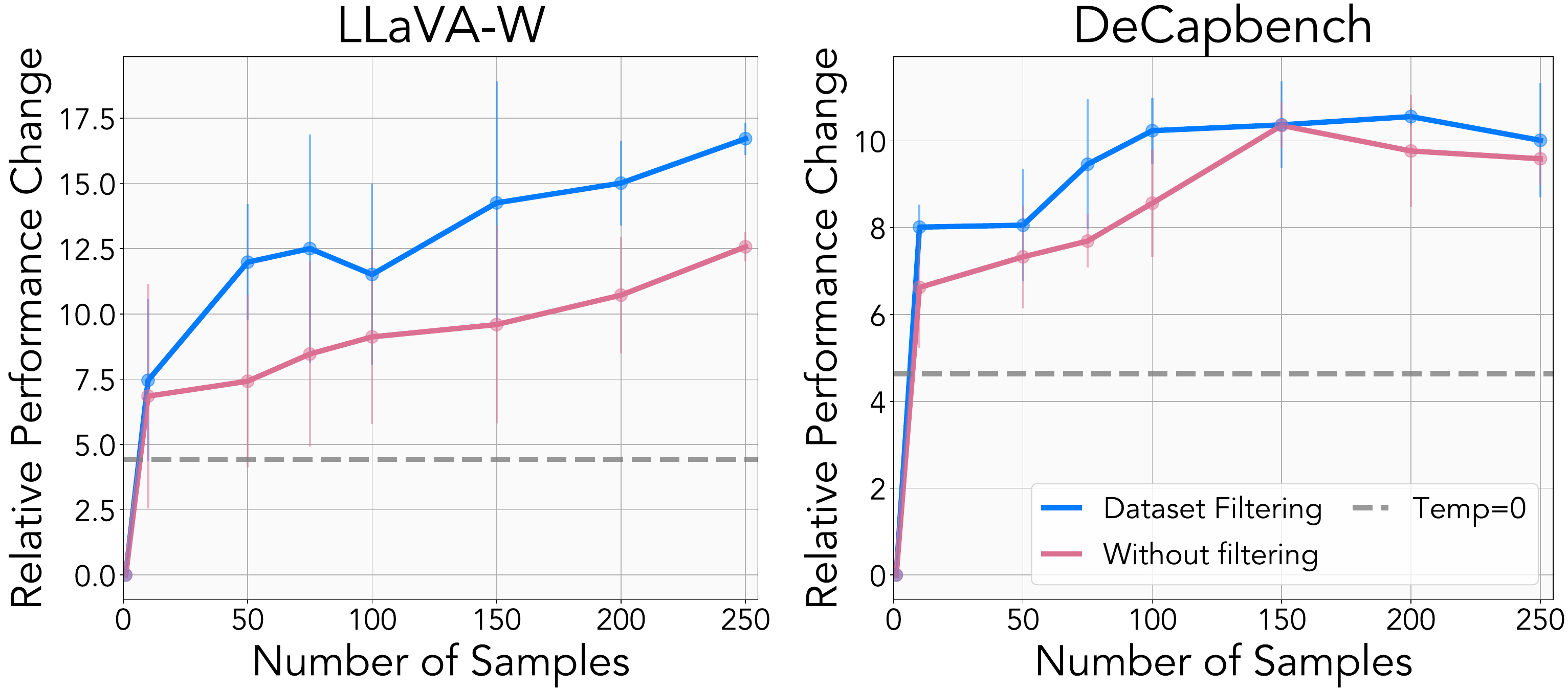}
  \caption{\textbf{Best-of-$N$ results with and without dataset filtering}. Filtering the dataset improves inference-time optimization by enabling better candidate selection during best-of-$N$ sampling.} 
  \vspace{-0.4cm}
  \label{fig:bon_filtering}
\end{figure}

\paragraph{Dataset Filtering} \label{appendix:data_filter}
Common strategies for filtering human preferences include: (1) removing duplicate entries, (2) filtering out cases where both responses are harmful or irrelevant~\cite{yeh2024reliable}, and (3) excluding low-margin examples where one response is only marginally better than the other~\cite{deng2025less}. Following these principles, we adopt a similar filtering strategy by removing duplicate captions, excluding examples where the reward difference is within a certain threshold, i.e., $|r_i-r_j|<\tau_\text{sim}$, and discarding comparison pairs where the preferred reward is below a threshold, i.e., $r_i <\tau_\text{neg}$. In practice, we use $\tau_\text{sim}=0.005$, $\tau_\text{neg}=0.7$ for DreamSim, and $\tau_\text{neg}=0.4$ for SBERT. In practice, training with dataset filtering leads to a small performance gain on alignment benchmarks and a bigger performance gap in Best-of-$N$ experiments as seen in Figure~\ref{fig:bon_filtering}.

\paragraph{Prompt Choice} \label{appendix:prompts}
To ensure that all image-to-text models can produce image descriptions to the best of their ability, we use the prompt recommended by the model distributor, as shown in Table~\ref{tab:prompt_detail}.

\begin{table}[h]
    \centering
    \resizebox{1.0\columnwidth}{!}{
    \begin{tabular}{l|c}
        \toprule
        Model & Prompt \\
        \midrule
        BLIP2 & ``this is a picture of'' \\
        LLaVA1.5 & ``Write a detailed description of the given image.'' \\
        LLaVA1.6 & ``Write a detailed description of the given image.'' \\
        LLaVA-OV & ``Write a detailed description of the given image.''  \\
        InternVL2 & ``Please describe the image in detail.'' \\
        \bottomrule
    \end{tabular}}
    \caption{Prompts used for image-to-text models.}\label{tab:prompt_detail}
    \vspace{-0.4cm}
\end{table}

\section{Model training details} \label{appendix:training_details}

\subsection{Reward Modeling} We use the AdamW optimizer~\cite{loshchilov2019decoupled} with a batch size of 2048 for 2 epochs. The learning rate is set to 3e-5 with a weight decay of 1e-4 for optimizing $\mathcal{L}_\mathrm{text}$, while $\mathcal{L}_\mathrm{img}$ and joint training use a learning rate of 2e-5 with no weight decay. We set $\lambda=1$ for joint training. Following the setup in~\cite{xu2024imagereward}, we fix 70\% of the transformer layers during training, which we found to outperform full fine-tuning. All models are trained using 8 H100 GPUs.

\subsection{DPO} We perform DPO to align Qwen-VL-Chat using our dataset CyclePrefDB-I2T. The model is trained for 5 epochs with the AdamW optimizer~\cite{loshchilov2017decoupled} and a weight decay of 0.05. We apply
a cosine learning rate schedule with a warmup ratio of 0.1 and a peak learning rate of $1\times10^{-5}$. Training is performed with a global batch size of 256. To enable more efficient training, we adopt LoRA tuning. The model is trained using 4 H100 GPUs.

\subsection{Diffusion-DPO} We use the Diffusion-DPO objective to align Stable Diffusion 1.5~\cite{rombach2022high} with preferences in our CyclePrefDB-T2I dataset. We use the AdamW optimizer~\cite{loshchilov2019decoupled} and train with an effective batch size of 512 (batch size 1 with 128 gradient accumulation steps on 4 H100 GPUs). We use learning rate $5\times10^{-8}$ and set $\beta=1000$ and train for $1500$ steps. Similarly to the Diffusion-DPO Pick-A-Pic model, we validate checkpoints with $380$ prompts from CyclePrefDB-T2I validation set and select the best checkpoint according to the mean alignment using the CycleReward-T2I reward model.

\begin{figure}[t]
  \centering
  \includegraphics[width=0.98\linewidth]{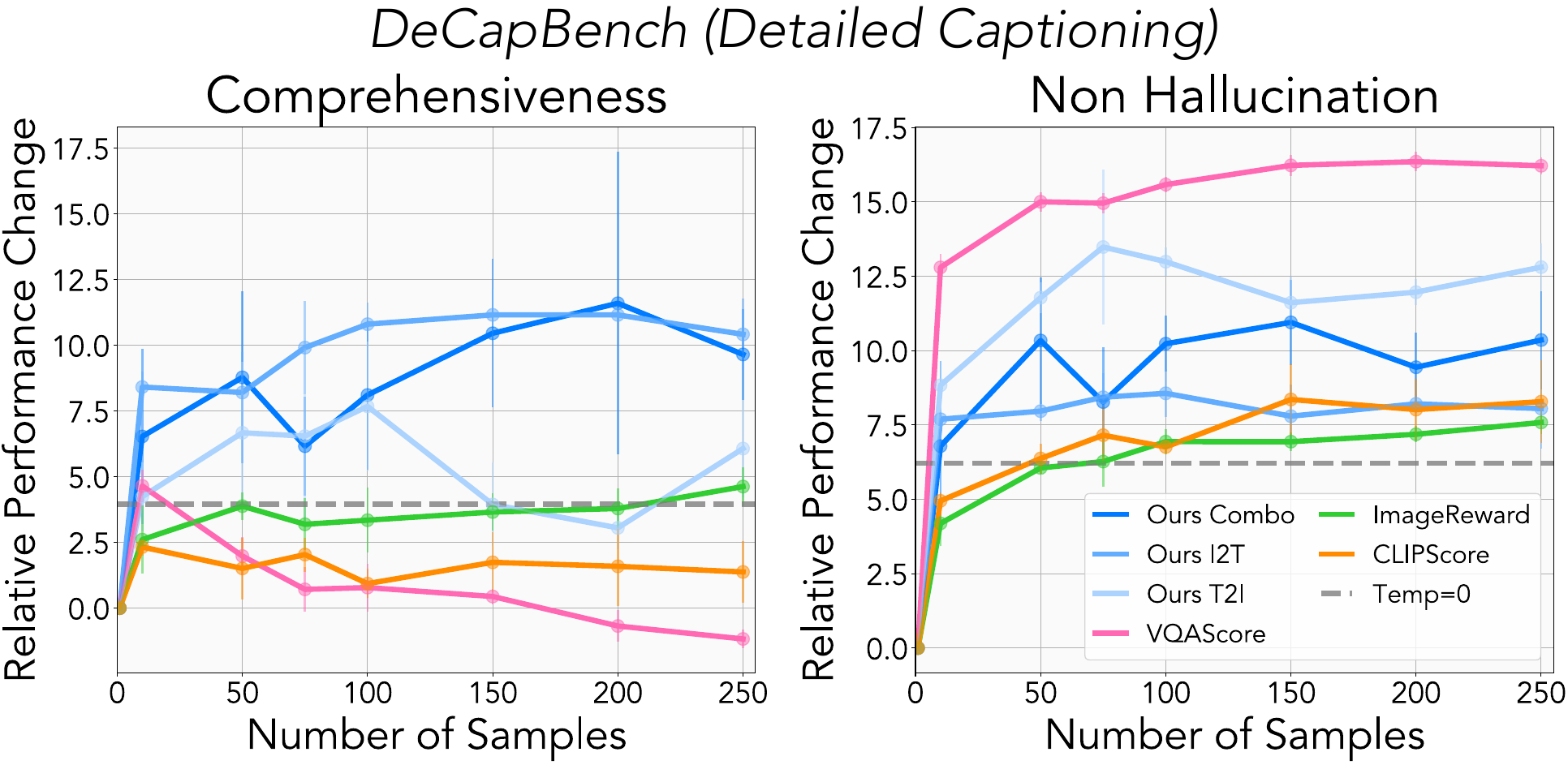}
  \vspace{-0.2cm}
  \caption{\textbf{DeCapBench Best-of-$N$ DCScore breakdown}. DeCapBench evaluation is performed with DCScore which combines scores for Comprehensiveness and Non-Hallucination in the left and right plots respectively.} 
  \vspace{-0.4cm}
  \label{fig:bon_decapbenchonly}
\end{figure}

\begin{figure*}[t]
  \centering
  \includegraphics[width=0.98\linewidth]{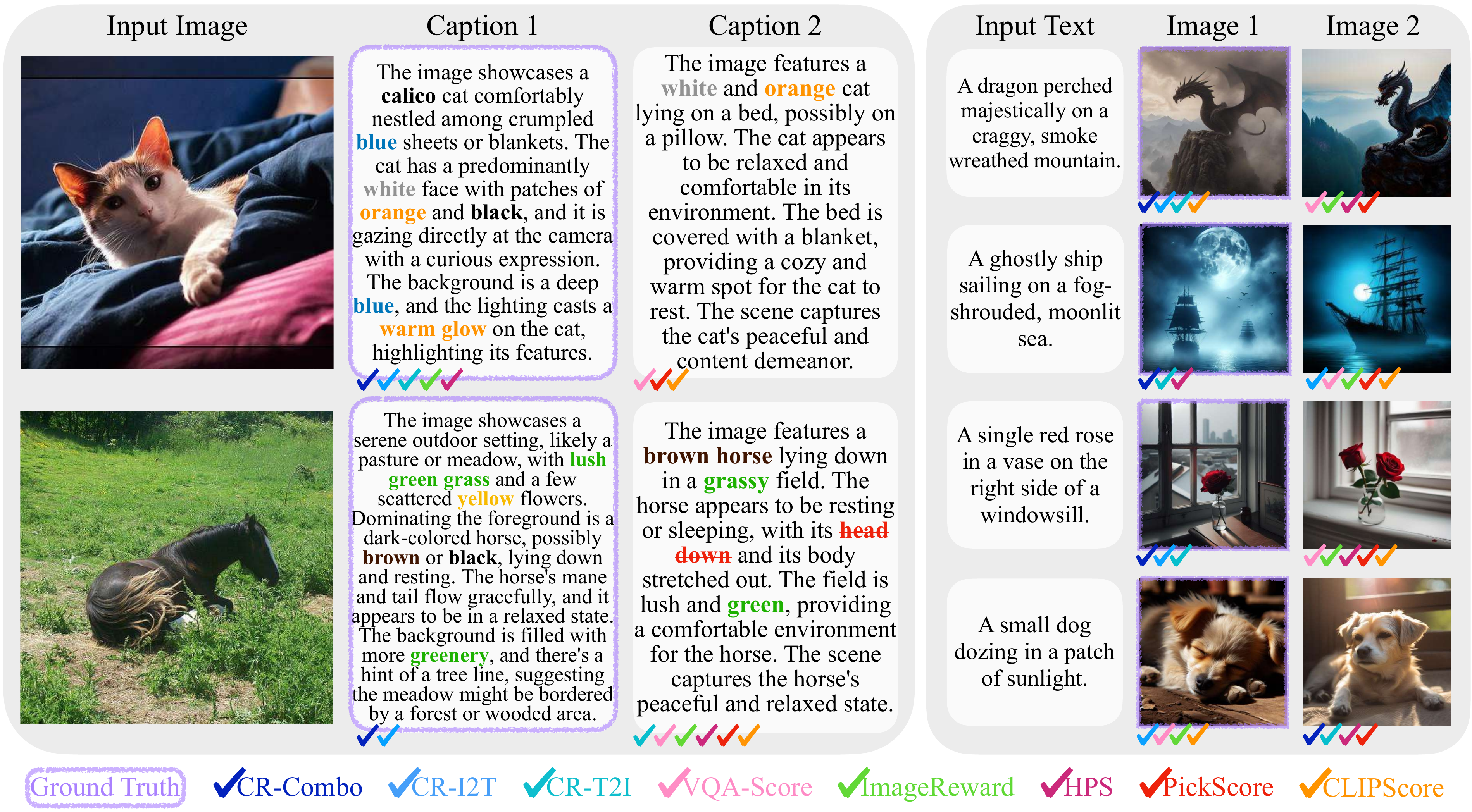}
  \vspace{-0.2cm}
  \caption{\textbf{Alignment metrics on DetailCaps-4870 and GenAI-Bench.} Our reward model excels at identifying detailed captions while performing competitively on GenAI-Bench. We also provide the ground truth label in purple.} 
  \label{fig:benchmark_examples}
\end{figure*}

\section{Additional Results}

\begin{figure*}[t]
  \centering
   \includegraphics[width=0.98\linewidth]{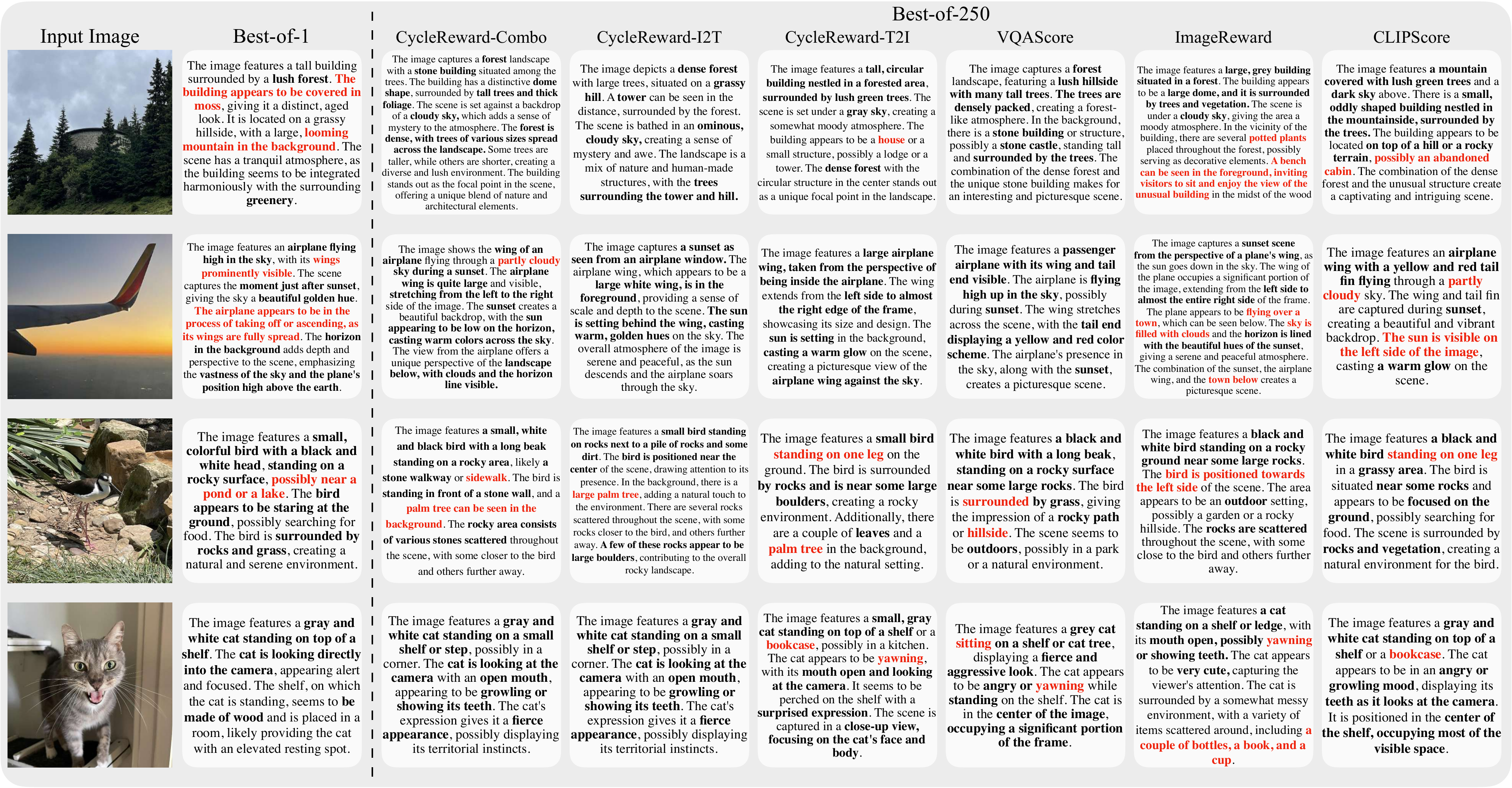}
    \caption{\textbf{Best-of-$N$ results on DeCapBench for different metrics.} Overall, our model increases the level of detail in captions while avoiding severe hallucinations.}
   \label{fig:bestof250}
\end{figure*}

\begin{figure*}[t]
  \centering
   \includegraphics[width=0.7\linewidth]{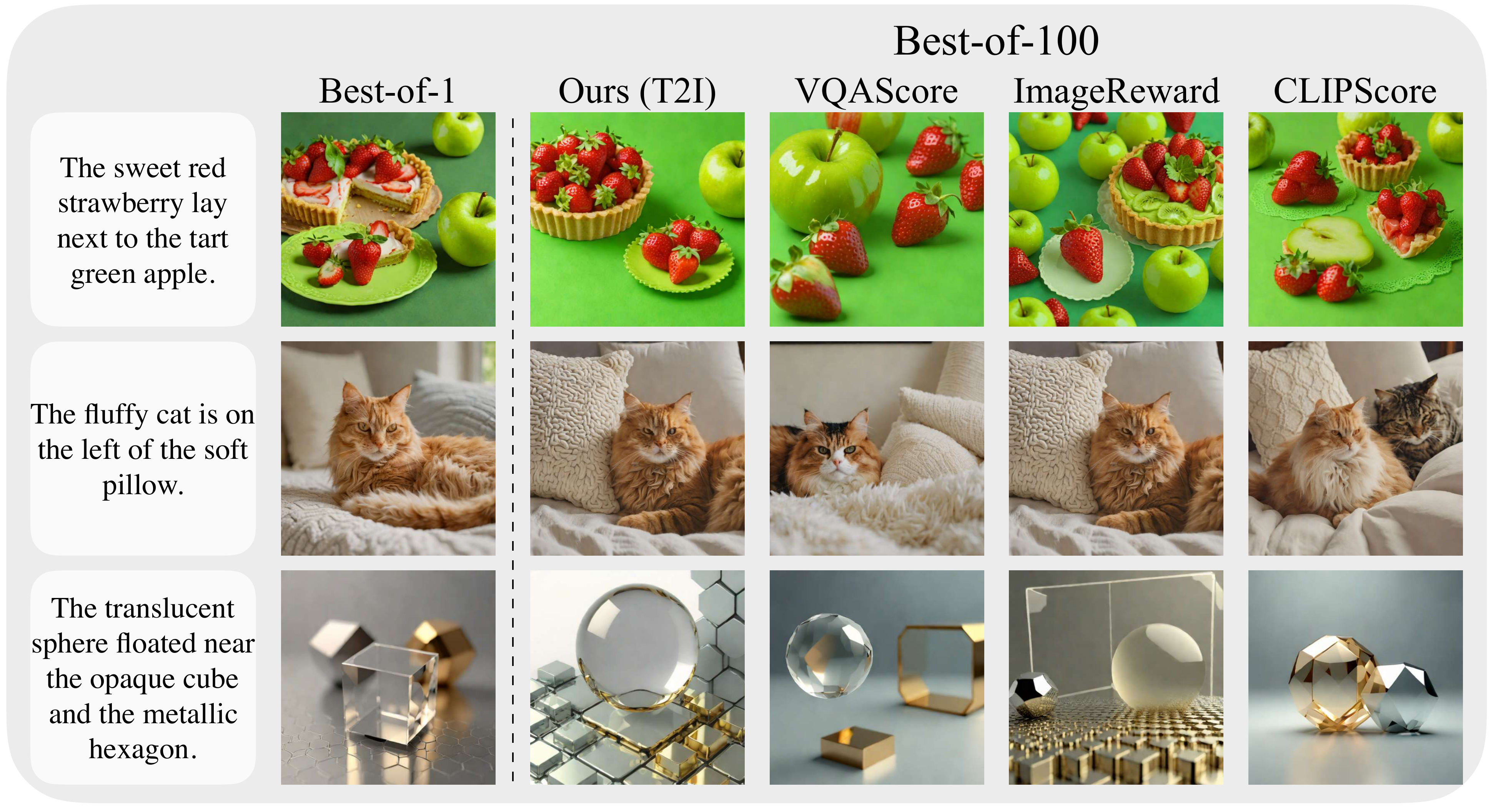}
   \caption{\textbf{Best-of-$N$ results on T2I-CompBench for different metrics.} Optimizing with our reward model generally improves results, while VQAScore excels at following positional relationships.}
    \label{fig:bestof100}
   \vspace{-0.4cm}
\end{figure*}

\begin{figure}[t]
  \centering
  \includegraphics[width=0.98\linewidth]{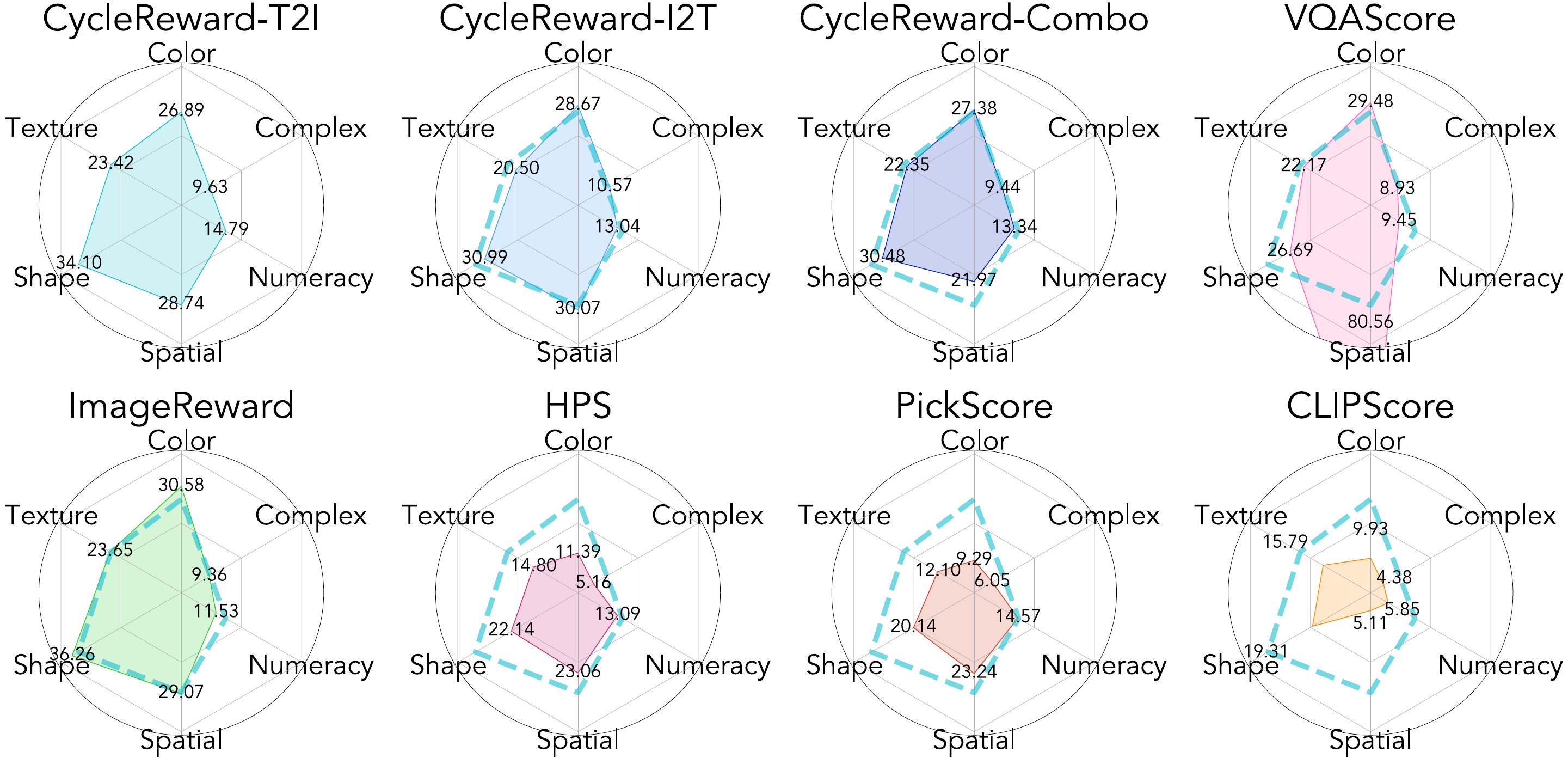}
  \vspace{-0.2cm}
  \caption{\textbf{Relative performance gain on T2I-CompBench from Best-of-$1$ to Best-of-$100$ across 6 categories}. We mark CycleReward-T2I's performance with a dashed line in all charts for comparison. While each metric has category-specific strengths, human-supervised ImageReward achieves the most balanced overall performance, followed closely by CycleReward-T2I.} 
  \vspace{-0.2cm}
  \label{fig:radars}
\end{figure}

\begin{figure}[t]
  \centering
  \includegraphics[width=0.98\linewidth]{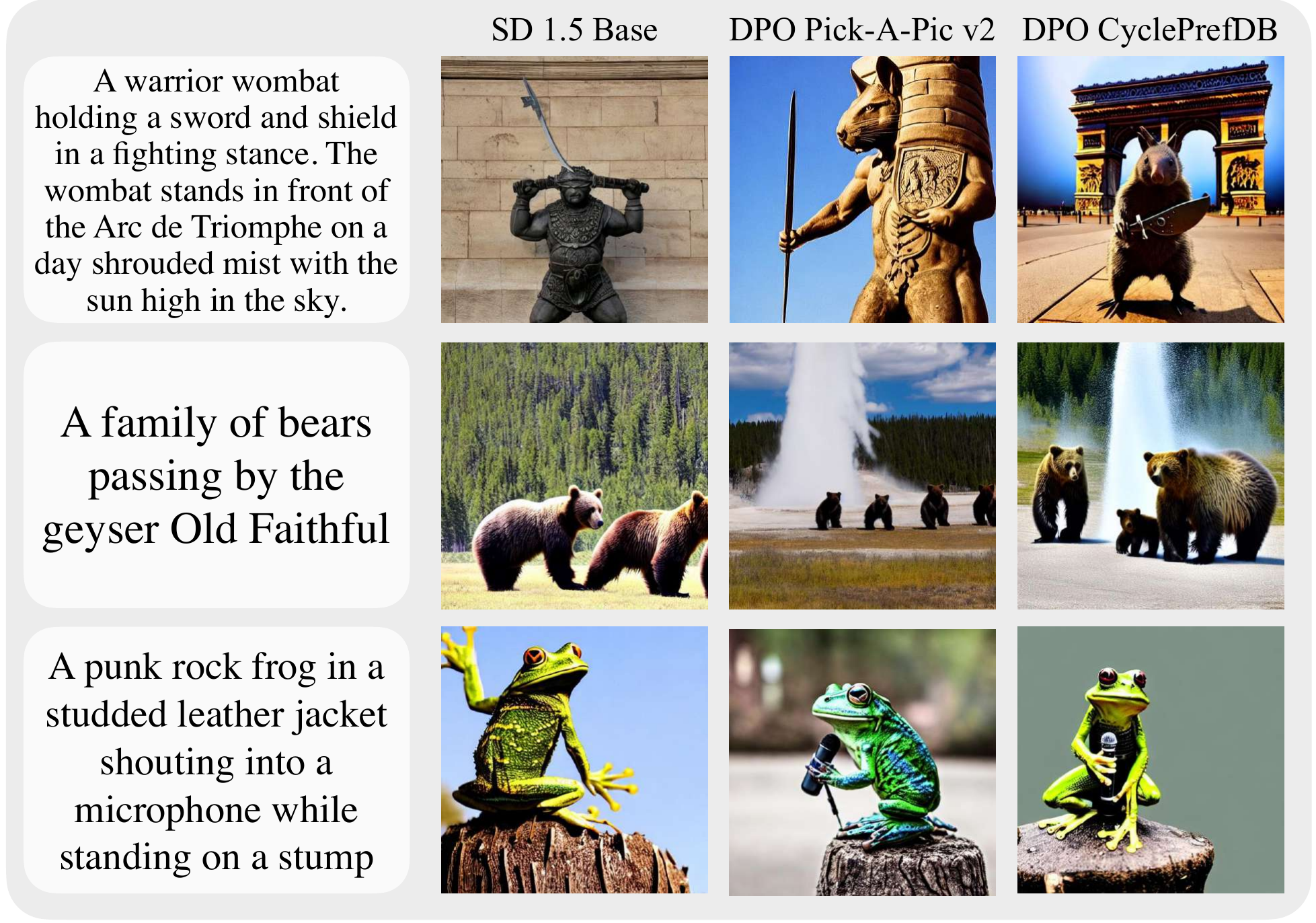}
  \vspace{-0.1cm}
  \caption{\textbf{Generated images from Diffusion DPO training.} We compare images generated by the base Stable Diffusion 1.5 model, a model trained on Pick-A-Pic v2, and a model trained on CyclePrefDB-T2I (ours). Our model captures complex visual details and often outperforms the Pick-A-Pic v2 model trained with human preferences.} 
  \vspace{-0.4cm}
  \label{fig:ddpo_t2i}
\end{figure}

\subsection{Alignment Metrics}\label{appendix:benchmark_examples}
Figure~\ref{fig:benchmark_examples} shows qualitative examples of CycleReward versus other alignment metrics with ground truth preferences in purple. Overall, our CycleReward (CR) models are more successful at assessing detailed captions while performing competitively on evaluating text-to-image generation.

\subsection{Best-of-N} \label{appendix:bon_cont}
Figures~\ref{fig:bestof250} and~\ref{fig:bestof100} show qualitative examples of how different metrics affect Best-of-N selection for detailed captioning and text-to-image generation, respectively. We show the initial (Best-of-1) output and compare it to the final output selected from the full candidate pool.

Figure~\ref{fig:bon_decapbenchonly} shows DeCapBench Best-of-N results separated into the Non-Hallucination and Comprehensiveness categories used by DCScore~\cite{yepainting} during evaluation. All CycleReward models lead to improvement in both categories, but CycleReward-Combo and CycleReward-I2T select the most comprehensive captions, while VQAScore and CycleReward-T2I yield the best non-hallucination scores. Note other metrics such as VQAScore and CLIP have tradeoffs which sacrifice description for accuracy.

\paragraph{Sampling Settings} To obtain candidate captions for Best-of-N sampling, we used a combination of temperature, nucleus, and prompt sampling with model LLaVA1.5-13B~\cite{liu2023llava,li2024llava}. We set temperature to $1.0$, top $p$ to $0.7$ respectively, and choose prompts randomly from the original LLaVA dataset prompts~\cite{liu2023llava}. Image candidates are generated using random seed sampling for diffusion models.

\paragraph{T2I-CompBench Categories} \label{appendix:t2icb}
Figure~\ref{fig:radars} shows Best-of-N results for individual categories in T2I-CompBench~\cite{huang2023t2i}. Our metric is most effective for complex prompts, whereas the VQAScore excels at spatial relationships. 

\begin{table}[t]
  \centering
  \resizebox{1.0\columnwidth}{!}{
  \begin{tabular}{l|ccc}
    \toprule
      & \multicolumn{3}{c}{Winoground} \\
    Method & Text Score & Image Score & Group Score \\
    \midrule
    \textit{Vision-language model}\\
    CLIPScore &  28.50	& 11.20	 & 8.25 \\
    VQAScore (3B) & 48.75 &	46.25 &	35.50\\
    VQAScore (11B) & \textbf{58.50} &	\textbf{56.25} & \textbf{44.75} \\
    \midrule
    \textit{Human preferences} \\
    HPSv2 &  26.75	& 10.50	& 8.25 \\
    PickScore & 23.75 &	12.50 &	6.75\\
    ImageReward & 43.00	 & 15.25 &	12.75 \\
    \midrule
    \textit{Cycle consistency} \\
    Raw Cycle Consistency & 29.00 & 17.50 & 13.50 \\
    \cellcolor{myapricot}CycleReward-T2I & \cellcolor{myapricot}40.00	& \cellcolor{myapricot}18.50 &	\cellcolor{myapricot}14.75 \\
    \cellcolor{myapricot}CycleReward-I2T & \cellcolor{myapricot}41.50 &	\cellcolor{myapricot}14.75 &	\cellcolor{myapricot}11.50 \\
    \cellcolor{myapricot}CycleReward-Combo & \cellcolor{myapricot}43.25	& \cellcolor{myapricot}16.75 &	\cellcolor{myapricot}13.25 \\
    \bottomrule
  \end{tabular}}
  \caption{\textbf{Winoground results.} Although we do not train on compositional reasoning tasks, CycleReward outperforms models trained on human preferences and raw cycle consistency. VQAScore, based on a large-scale VLM, outperforms all other metrics.}
  \vspace{-0.3cm}
  \label{tab:winoground}
\end{table}

\subsection{Winoground} \label{appendix:winoground}
We use the Winoground dataset to benchmark performance on visio-linguistic compositional reasoning in Table~\ref{tab:winoground}. Winoground comprises $400$ examples, each containing two image-text pairs where the texts use the same words in different orders to convey different meanings. Performance is measured by how often a metric matches the correct image with its corresponding text. Surprisingly, CycleReward variants, trained solely on self-supervised rewards, outperform all metrics trained on expert human annotations. All CycleReward variants are better at selecting text for an image (text score) than selecting images from a given description (image score). While our method outperforms CLIPScore and raw cycle consistency, VQAScore outperforms all other metrics. Note that VQAScore benefits from LLM scale (x6 and x24 larger than other methods). Additionally, our model is trained on visual descriptions instead of reasoning tasks, unlike the CLIP-FlanT5 model used in VQAScore.

\subsection{More Ablations} \label{appendix:more_ablations}

We study additional ablations on CycleReward-I2T trained on image-to-text comparison pairs. (1) \textit{Objective Function}: We apply MSE loss to directly regress the cycle consistency score. Surprisingly, this results in a severe performance drop. We hypothesize that Bradley-Terry loss~\cite{ouyang2022training,stiennon2020learning} better captures relative preferences effectively, while MSE focuses on regressing exact score values. (2) \textit{Dataset Size}: We maintain all configurations but train on a subset of DCI 1K images. The performance gap highlights the efficacy of scaling our dataset. (3) \textit{Dataset Filtering}: We train a model without dataset filtering, which causes a small performance drop on alignment evaluation, with a larger decrease for Best-of-$N$ selection (Appendix~\ref{appendix:data_filter}). We believe discarding noisy comparison pairs helps select better candidates as the sample pool expands.

\begin{table}[h] 
  \centering
  \resizebox{1.0\columnwidth}{!}{
  \begin{tabular}{lcc}
    \toprule
    Ablation & DetailCaps-4870 & GenAI-Bench \\
    \midrule
    \cellcolor{mygray}Best variant (CR-I2T)& \cellcolor{mygray}\textbf{58.02} & \cellcolor{mygray}\textbf{53.49}\\
    MSE loss & 41.87& 40.57\\
    1K images & 52.86 & 44.39\\
    Without filtering & 57.28  & 51.92\\
    \bottomrule
  \end{tabular}}
  \vspace{-0.2cm}
  \caption{\textbf{Effect of objective function, data size, and filtering.} Choices used by our model are in \colorbox{mygray}{gray}.}
  \vspace{-0.3cm}
  \label{tab:ablation}
\end{table}

\subsection{DPO}
Figure~\ref{fig:ddpo_t2i} shows comparisons between the base Stable Diffusion 1.5 model, the Diffusion-DPO model trained with Pick-a-Pic v2, and the Diffusion DPO model trained with our CyclePrefDB-T2I dataset. Training with cycle consistency preferences achieves comparable results as training with Pick-a-Pic v2, despite lacking human labels. Furthermore, our dataset is about half the size of Pick-a-Pic v2.

\subsection{Failure Cases} \label{appendix:failures}

\begin{figure}[t]
  \centering
  \includegraphics[width=1\linewidth]{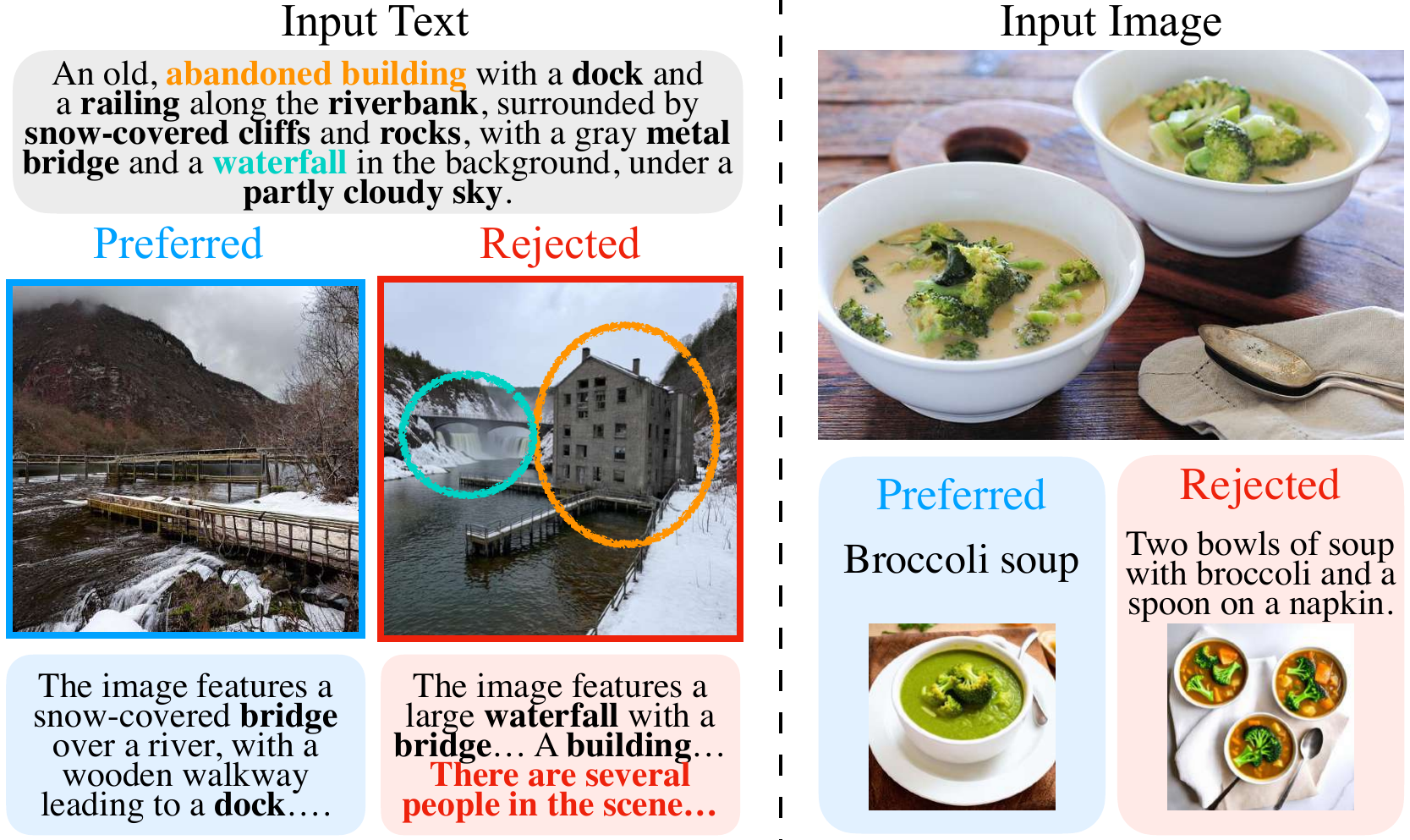}
  \caption{\textbf{Failure cases.} (Left): Despite being faithful to the input text, the right image is rejected as the reconstructed text contains \textcolor{red}{hallucinations} inconsistent with the original prompt. (Right): The short caption is preferred over the descriptive caption due to an error in text-to-image generation. Under each caption we show the corresponding reconstructed images.}
  \vspace{-0.4cm}
  \label{fig:failures}
\end{figure}

Although we propose cycle consistency as a self-supervised signal for learning image-text alignment, our method has several limitations. A common source of failure is poor reconstructions which mislead preferences determined by cycle consistency seen in Figure~\ref{fig:failures}. Our method also inherits biases from the underlying models used for reconstructions and and similarity measurements. Stable Diffusion 3 has a 77-token limit which limits consideration of longer texts, and LLaVA-1.5-3B can be prone to hallucinations. DreamSim often favors images with similar foregrounds over backgrounds~\cite{fu2023dreamsim}, and SBERT is sensitive to text style. Furthermore, we observe worse text-to-image performance, which may partially stem from dataset differences. HPSv2, PickScore, and ImageReward are trained on prompts from real users often describing artwork, whereas CycleReward is trained on LLM-summarized descriptions for natural images. Moreover, cycle consistency primarily considers preservation of information, while other aspects such as aesthetics or style may also affect human preferences. Future work could address these challenges by improving reconstruction quality, prompt diversity, and applying cycle consistency in different scenarios.

\begin{figure*}[t]
  \centering
  \includegraphics[width=0.98\linewidth]{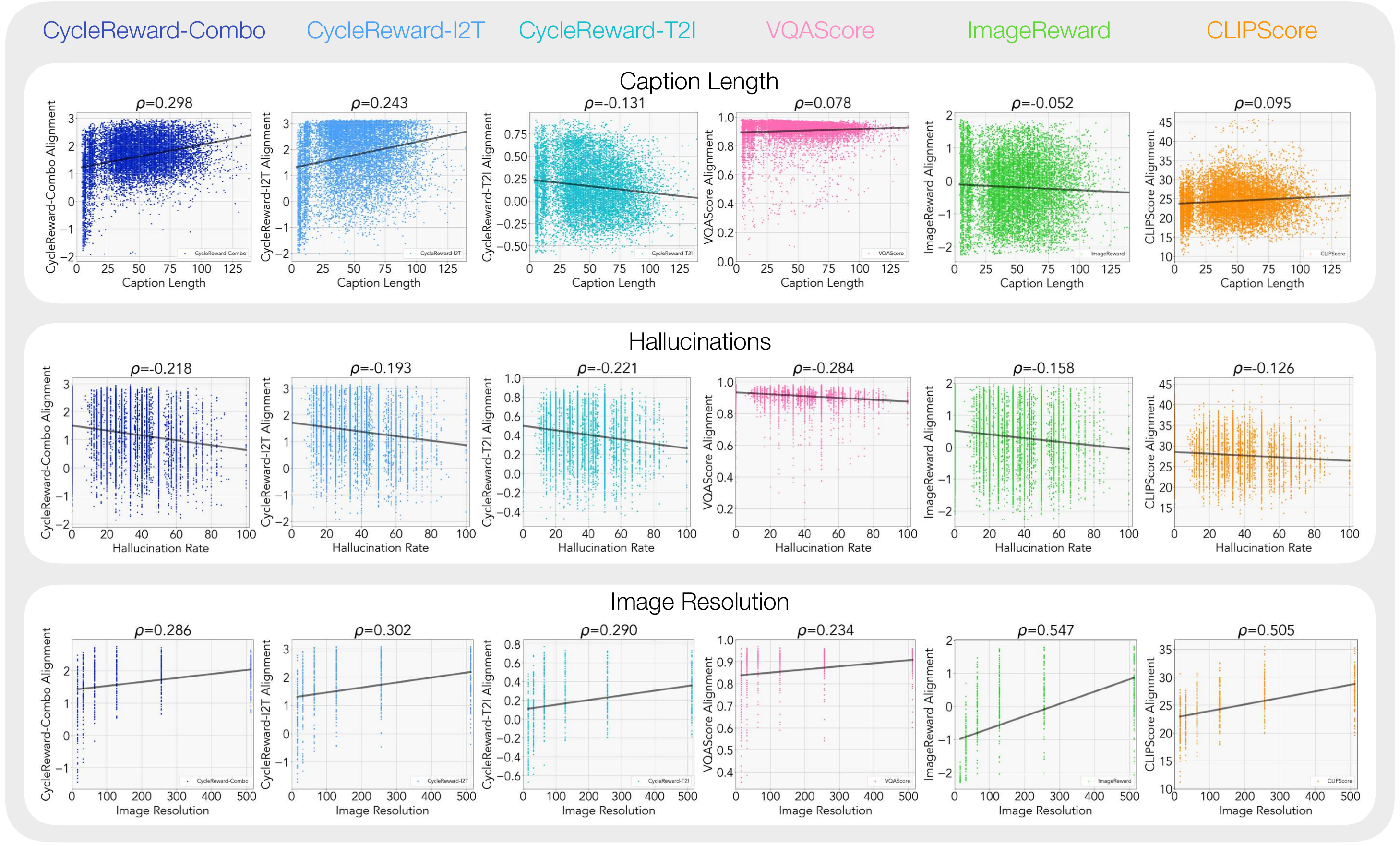}
  \vspace{-0.2cm}
  \caption{\textbf{Text and image data trends for different alignment metrics.} For each metric, we plot how various factors (shown on each row) affect alignment scores. Note that different alignment measurements are not comparable by scale, but their correlation with each specific factor can be measured. CycleReward-I2T and CycleReward-Combo tend to prefer longer captions, while models trained with text-to-image comparison pairs (ImageReward and CycleReward-T2I) generally prefer shorter captions. In terms of number of hallucinations and image operations, we find that all metrics show consistent correlation directions, albeit some metrics such as VQAScore and CycleReward exhibit greater sensitivity to text inaccuracies.}
  \vspace{-0.4cm}
  \label{fig:alignment_chart}
\end{figure*}

\section{Reward Model Trends}
We investigate how text and image properties affect different metrics' alignment preferences for the following factors: caption density, object hallucination, image density, and resolution in Figure~\ref{fig:alignment_chart}. For each specific factor, we plot the alignment score for individual image, text pairs based on the relevant image or text characteristic. The title of each plot reports the Pearson correlation coefficient between the alignment score and respective factor. We also display the line of best fit. Note that the scale and range of alignment scores are different and therefore not directly comparable between metrics. Because of this we instead focus on overall trends and correlations between each factor and alignment.

\paragraph{Caption Length}
To examine which reward models generally prefer long or short captions, we first create a dataset of images paired with captions of various lengths. We utilize the test and validation sets of the DCI~\cite{urbanek2024picture} dataset for this task, where each image is paired with a long, descripitive text. For each image, we use an LLM (Meta-Llama-3.1-8B-Instruct~\citep{dubey2024llama}) to create captions of different lengths but asking for summaries with different numbers of words, similarly to Huh et al.~\cite{huh2024prh}. We ask for summaries of lengths 5, 10, 20, ..., 100 words, and sample 5 different captions for each length with temperature 0.6 and top p 0.9. This results in 11241 unique image, caption pairs after eliminating duplicates and removing ``here is a summary`` text. 

In Figure~\ref{fig:alignment_chart} (top row), we plot the alignment trend for different metrics versus caption length. The Pearson correlation coefficient $\rho$ is reported at the top of each plot. Because captions can be informative or contain mistakes regardless of their lengths, we expect these plots to be noisy. All methods, except for CycleReward-T2I and ImageReward, have positive Pearson Correlation coefficients - meaning they in general longer captions are preferred. However the correlation between caption length and alignment is much weaker for VQAScore and CLIP compared to CycleReward-Combo and CycleReward-I2T.

\paragraph{Hallucination Rate}
To view how hallucinations affect alignment preferences, we use the M-HalDetect dataset~\cite{gunjal2024detecting}. This dataset contains images paired with captions from InstructBLIP~\cite{instructblip}. We use the validation and training sets for this dataset totaling 14143 image caption pairs. Each caption is divided into sections which have been annotated for their accuracy and having hallucinations. We compute the fraction of hallucinated parts in each caption and plot this value against the alignment in Figure~\ref{fig:alignment_chart} (middle row). All metrics tend to prefer captions with less hallucinations (lower hallucination rate), although with different correlation strengths - VQAScore having the strongest correlation followed by CycleReward-T2I and CycleReward-Combo.

\paragraph{Image Resolution}
For text-to-image, we examine how images of different resolutions affect alignment with the text. To this end, we gather 100 ``upsampled'' text descriptions created by prompting GPT-4o\cite{gpt-4o} to add details to short captions from MSCOCO~\cite{lin2014microsoft}. Text descriptions are encouraged to be visually informative and no longer than 77 tokens. We use SDXL~\cite{rombach2022high} to generate images for each text description at 512$\times$512 resolution. We resize the images to resolutions 256, 128, 64, 32, 16 and compute alignment at each stage in Figure~\ref{fig:alignment_chart}(bottom row). For all metrics, alignment is generally not affected when resizing from 512 to 256 and 128 pixels, and then drops off steeply as the resolution goes from 64 to 16. Note that CycleReward and ImageReward preprocess images to be size  while CLIP and VQAScore preprocessing resizes image to 336.

\end{document}